%% file: TDSC_EDIT_Full_Version.tex
\documentclass[10pt,journal,compsoc]{IEEEtran}
 \usepackage{multicol, blindtext} % Tables
 \usepackage{hyperref}
 \usepackage{multirow}  
 \usepackage{graphicx,caption,subcaption}
 \usepackage{soul}
 \usepackage[utf8]{inputenc}
 \DeclareCaptionLabelFormat{AppendixTables}{Appendix \Alph{table}}
 \usepackage{afterpage}  % load the afterpage package
 \usepackage{longtable}  % load the longtable package
 \usepackage{tcolorbox}
 \usepackage{xcolor,colortbl}
\usepackage[ruled]{algorithm2e} % For algorithms

\newcommand{\updated}[1]{\textcolor{black}{#1}}
\newcommand{\new}[1]{\textcolor{black}{#1}}

\renewcommand{\small}{\fontsize{9.5pt}{7pt}\selectfont}

\ifCLASSOPTIONcompsoc
  \usepackage[noadjust]{cite}
\else
  \usepackage{cite}
\fi
\ifCLASSINFOpdf
\else
\fi
\usepackage{ragged2e}
\usepackage{url}
\renewcommand{\footnotesize}{\fontsize{6pt}{7pt}\selectfont}
\usepackage{fancyhdr}
\usepackage{cite}
\usepackage{amsmath,amssymb,amsfonts}

\usepackage{graphicx}
\usepackage{textcomp}
\usepackage{xcolor}

\usepackage{nicefrac}
\usepackage{siunitx}
\usepackage{array,framed}
\usepackage{booktabs}
\usepackage{
  color,
  float,
  epsfig,
  wrapfig,
  graphics,
  graphicx,
  subcaption
}
\usepackage{textcomp,amssymb}
\usepackage{setspace}
\usepackage{latexsym,fancyhdr,url}
\usepackage{enumerate}
\usepackage{tcolorbox}
\usepackage{soul}
\usepackage{algorithm2e}
\usepackage{algpseudocode}
\usepackage{graphics}
\usepackage{xparse} % argument parsing -- \edist
\usepackage{xspace}
\usepackage{multirow}
\usepackage{csvsimple}
\usepackage{tabularx}
\usepackage{balance}
\newcolumntype{C}[1]{>{\centering\arraybackslash}m{#1}}

\begin{document}
%
% paper title
% Titles are generally capitalized except for words such as a, an, and, as,
% at, but, by, for, in, nor, of, on, or, the, to and up, which are usually
% not capitalized unless they are the first or last word of the title.
% Linebreaks \\ can be used within to get better formatting as desired.
% Do not put math or special symbols in the title.
\title{Alert-ME: An Explainability-Driven Defense Against Adversarial Examples in Transformer-Based Text Classification}

\author{\IEEEauthorblockN{Bushra Sabir\IEEEauthorrefmark{1,2,3},
Yansong Gao\IEEEauthorrefmark{4}, 
Alsharif Abuadbba \IEEEauthorrefmark{1}, 
M. Ali Babar\IEEEauthorrefmark{2,3}
}

\IEEEauthorblockA{\IEEEauthorrefmark{1} CSIROs Data61.}
\IEEEauthorblockA{\IEEEauthorrefmark{2} School of Computer Science, The University of Adelaide.}
\IEEEauthorblockA{\IEEEauthorrefmark{3} CREST - The Centre for Research on Engineering Software Technologies.}
\IEEEauthorblockA{\IEEEauthorrefmark{4} Department of Computer Science and Software Engineering, The University of Western Australia.}
% <-this % stops an unwanted space
}
\IEEEtitleabstractindextext{%
\begin{abstract}
\justifying Transformer-based Text Classifiers (TTC), exemplified by models like BERT, Roberta, T5, and GPT, have showcased impressive proficiency in Natural Language Processing (NLP). Nonetheless, their vulnerability to adversarial examples (AEs) presents a substantial security concern. Existing efforts to fortify model robustness either entail high computational costs or lack transparency and interpretability.
This paper introduces a holistic framework called \underline{E}xplainability-driven \underline{D}etection, \underline{I}dentification, and \underline{T}ransformation (EDIT) to address the shortcomings of prevailing inference-time defenses. (i) EDIT employs explainability methodologies (e.g., attention maps, integrated gradients) and frequency features for automated detection and  identification of perturbed words while providing insights into the model's decision-making process. (ii) Building upon detection, EDIT purifies detected adversarial examples through an optimal transformation mechanism that utilizes pre-trained embeddings and model feedback to find the best replacements for identified perturbed words. (iii) To integrate human intelligence, the framework automates the process and triggers alerts for a security analyst to ensure secure safeguards and enable proactive human feedback.
We comprehensively evaluate our framework on BERT and ROBERTA, trained on four SOTA text classification datasets: IMDB, YELP, AGNEWS, and SST2. We also assess our framework against seven SOTA Word Substitution Attacks (WSA) at character, word, and multi-levels of granularity. The results show that EDIT achieves an average F-score and balanced accuracy of 89.69\% and 89.70\%, respectively, across all datasets and classifiers.
Compared to four SOTA inference defenses, our detector is 1.22 times better in balanced accuracy (BAL\_ACC) and 1.33 times better in F1-Score. It also excels in operational efficiency, being about 83 times faster in feature extraction. The identification module outperforms techniques like ReplaceScore, FreqScore, and ExplainScore, achieving 1.125 times better BAL\_ACC and Recall, 1.142 times higher AUC, 1.120 times greater Precision, and 1.149 times improved F1-Score. It significantly reduces false positives and negatives by approximately 1.537 and 1.172 times, respectively.
Our transformation module effectively converts adversarial examples into non-adversarial counterparts with an accuracy of 91\%. EDIT identifies the necessity for human intervention with an average median accuracy of 89\%. With alerts integrated, EDIT achieves an average median accuracy of 90\% across all evaluated datasets, models, and types of attacks, all while maintaining computational efficiency, averaging  6.94 seconds.
\end{abstract}

}

% make the title area
\maketitle

% To allow for easy dual compilation without having to reenter the
% abstract/keywords data, the \IEEEtitleabstractindextext text will
% not be used in maketitle, but will appear (i.e., to be "transported")
% here as \IEEEdisplaynontitleabstractindextext when the compsoc 
% or transmag modes are not selected <OR> if conference mode is selected 
% - because all conference papers position the abstract like regular
% papers do.
\IEEEdisplaynontitleabstractindextext
% \IEEEdisplaynontitleabstractindextext has no effect when using
% compsoc or transmag under a non-conference mode.

% For peer review papers, you can put extra information on the cover
% page as needed:
% \ifCLASSOPTIONpeerreview
% \begin{center} \bfseries EDICS Category: 3-BBND \end{center}
% \fi
%
% For peerreview papers, this IEEEtran command inserts a page break and
% creates the second title. It will be ignored for other modes.
\IEEEpeerreviewmaketitle

%786
\IEEEraisesectionheading{\section{Introduction}
\label{introduction}}
\IEEEPARstart{T}
ransformer-based text classifiers, including BERT, Roberta, T5, and GPT-3, have revolutionized Natural Language Processing (NLP) with their exceptional performance and contextual understanding capabilities \cite{vaswani2017attention,devlin2018bert,liu2019roberta}.  These models are the current State-Of-The-Art (SOTA) and have been leading the leader-board for many tasks, including detecting, rating, and classifying user reviews, hate speech, toxic content, cyber-attacks, and news on online platforms \cite{vaswani2017attention,devlin2018bert,liu2019roberta}. Big technology giants such as Google \footnote{https://www.perspectiveapi.com/}, Facebook \footnote{https://ai.facebook.com/blog/cs2t-a-novel-model-for-hate-speech-detection/}, and Amazon \footnote{https://aws.amazon.com/comprehend/} have been using Transformer-based models for text moderation on their platforms.  

However, these models, like their other Deep Neural Network (DNN) counterparts, are not foolproof and are vulnerable to carefully designed Adversarial Examples (AEs) \cite{li2020bert,jin2019bert}.
AEs are generated by perturbing the original example in a way that retains its original semantics but induces misclassification at test time, resulting in evading the target text classifier. 
For example, transforming ``The Fish N Chips are excellent'' to ``The Fish N Chips are first-class'' changes the target model output (BERT for restaurant review classification) from a positive to a negative restaurant review.
Adversaries utilize several obfuscation methods to generate AEs ranging from substituting to inserting critical words (e.g., HotFlip \cite{ebrahimi-etal-2018-hotflip}, TextFooler \cite{jin2019bert}, characters
(e.g., DeepWordBug \cite{gao2018black}, TextBugger \cite{li2018textbugger}),  phrases  (e.g., \cite{xu2021grey,le2020malcom}) in the text to fool the target model. 

\textbf{Motivation. } 
Researchers have recently explored various defence strategies to fortify NLP models against evasion attacks. These defences can be broadly categorized into two main types: training-time defences and inference-time defences. Training-time defences operate by modifying the training data or training process itself, with the objective of enhancing the model's resilience. This may involve exposing the model to AEs during training or encoding input data in a manner that supports robustness. Notable examples of these defences include adversarial training \cite{wang2020infobert}, robust encoding \cite{jones2020robust}, and semantic encoding \cite{wang2021natural}.  
Note that these defences often require retraining, rendering them computationally and resource-intensive \cite{le2022shield,MLMTAED2023}.
%update  \garrison{mentioning interpretability as a challenge, computationally less expensive }

In contrast, inference-time defense mechanisms such as MLMD \cite{MLMTAED2023}, WDR \cite{mosca2022suspicious}, RDE \cite{yoo2022detection}, and SHAP \cite{huber2022detecting} provide plug-and-play capabilities specifically for detecting adversarial samples during the inference phase. These mechanisms eliminate the need for retraining the victim model, thereby enhancing adaptability and efficiency. However, the primary focus of these studies \cite{MLMTAED2023, mosca2022suspicious, yoo2022detection, huber2022detecting} is limited to detecting and blocking adversarial examples (AEs), without providing insights into model vulnerabilities or offering solutions to correct these examples.
While few initiatives like Textshield \cite{shen2023textshield} and DISP \cite{zhou2019learning} focused on both AE detection and purification to clean examples (CE), they utilized threshold-based static methods for perturbed word identification and replacement to purify AE. For example, Textshield \cite{shen2023textshield} consider words with  explainability score $> \beta$ as perturbed words and replace them with their most frequent synonyms from NLTK \cite{bird2006nltk}. This limits their adaptability to evolving linguistic contexts, potentially resulting in sub-optimal identification and replacements.
Furthermore, these approach rely on less interpretable features, such as outputs from the last layer or last attention layer, presents another significant drawback \cite{MLMTAED2023, mosca2022suspicious, yoo2022detection}. This lack of transparency hinders the ability of security analysts to understand and trace the decision-making process of these models, complicating efforts to diagnose and remedy vulnerabilities effectively. 
Although some recent studies \cite{shen2023textshield, huber2022detecting} have begun to explore more interpretable features for AE detection using advanced techniques like SHAP \cite{lundberg2017unifiedSHAP}, Layerwise Relevance Propagation (LRP) \cite{montavon2019layer}, and Integrated Gradients (IG) \cite{nauta2023anecdotal}, these methods are computationally demanding (albeit less than retraining the model from scratch). Particularly when applied to complex models such as BERT and RoBERTa, the computational overhead makes these techniques less feasible for real-time applications \cite{chefer2021generic}.
 Lastly, existing methods lack a mechanism to alert or provide heuristics about the AEs detection or correction to the security analyst.  The absence of such mechanisms significantly impedes timely intervention and comprehensive vulnerability assessment, crucial for reinforcing the resilience of TTC models against adversarial attacks.
%, which are critical for reinforcing model resilience against adversarial attacks.

In summary, existing defenses against adversarial evasion attacks in NLP models are either computationally intensive or rely on limited methods, lacking interpretability and analyst alert mechanisms. Hence, there's a critical need for a comprehensive pipeline that defends TTC models through automated, lightweight, and optimized processes for detecting, identifying, and transforming adversarial examples.

\textbf{Challenges. } To defend TTC against evasion attacks, we address three major challenges:
\texttt{C1. How can we leverage explainability for TTC to train an accurate adversarial detector?}
The complex architecture of TTC, with self-attention layers, skip connections, and non-linearities, complicates traditional explainability measures like SHAP and LRP. This complexity hinders understanding why TTC models misclassify AEs, making it challenging to distinguish between clean and adversarial instances.
\texttt{Solution}: We efficiently achieve explainability by leveraging attention across each layer and integrating gradients across heads to generate attention maps. This is computationally efficient due to the direct use of attention weights and streamlined computation of attention maps through gradient integration. Additionally, we consider the distribution of attention maps over all tokens, extracting statistical features, and integrate frequency-based measures to improve detection.
\texttt{C2. How can we accurately identify perturbed words and optimally transform them to convert AEs to clean examples (CEs)?}
Transformer models' complexity makes identifying and transforming perturbed words challenging. Evasion attacks often involve perturbing multiple words to create natural-looking AEs, exploiting language ambiguity and text data's high-dimensional nature.
\texttt{Solution:} We use attention-based relevance scores and outlier detection to identify perturbed words. Multiple replacement options (synonym, semantic, contextual) are identified, and our algorithm purifies detected AEs using pre-trained embeddings and model feedback for optimal replacements.
\texttt{C3. How to identify cases where human intervention is required?}
Automated defenses may fail to distinguish certain adversarial instances due to concept drift and evolving threats. Human oversight is essential in critical fields like cybersecurity and healthcare.
\texttt{
Solution:} We use adversarial detectors and TTC feedback to identify cases requiring human intervention, enhancing reliability and addressing uncertainties that automated defenses might miss.

EDIT addresses existing inference-time defense shortcomings by employing computationally efficient explainability features for automated AE detection. It dynamically identifies optimal replacements from a diverse pool, handling various levels of granularity. Moreover, it proactively alerts analysts and offers threat intelligence information.

\textbf{Contribution. } Our main contributions are summarized as follows:

\begin{itemize}
\item We build a meta-model, namely Adversarial Detector ($M_\mathrm{ad}$) 
by extracting features from TTC attention maps, integrated gradients, and frequency distribution to effectively differentiate between AEs and CEs.
\item We devise an automated approach that integrates attention, frequency, and model feedback to identify perturbed words accurately.
\item We present an optimal transformation that dynamically find a suitable replacements for an identified perturbed word from a pool of replacement options, converting or purifying identified AEs into non-adversarial forms. Additionally, the EDIT intelligently identify specific scenarios that require human intervention.
\item Lastly, we comprehensively evaluate our proposed detector, our identification and transformation module over four datasets, two TTCs (BERT and ROBERTA) and seven SOTA adversarial attacks (4*2*7=56 settings). Our EDIT framework provides valuable threat intelligence and actionable insights to security analysts facilitating human-AI collaboration. These insights can enhance the robustness of the defence mechanism and identify vulnerabilities in the target model. We have made our code available for reproduciblity and comparison \footnote{https://shorturl.at/LyvLz}. 
\end{itemize}

\section{Preliminaries and Background}
\label{sec:relwork}
% In~\secref{sec:intro} 
The following section aims to provide a comprehensive overview of the relevant background in the field.

\subsection{Transformer-based Text Classification Models. }
The transformer architecture was first introduced by Vaswan \textit{et al.} \cite{vaswani2017attention}, and has since become a popular choice for NLP tasks.
The transformer-based text classifier $T_m$ is a machine learning algorithm that takes as input a sequence $X$ of word embeddings and goes through multiple Transformer layers to output a probability distribution over the possible class labels $y$. Each layer includes a self-attention sub-layer that computes self-attention heads and linearly projects their concatenated outputs. Additionally, the model incorporates a residual connection, layer normalization, and feed-forward sub-layer. The final output representation $O_N$ undergoes linear projection and softmax function. During training, the model optimizes cross-entropy loss using gradient information to minimize disparities with true class labels.

\subsection{Prediction Confidence Score (PCS). } PCS($x$,$y$) \cite{sabir2021reinforcebug} of model $F$ depicts the likelihood of an input $x \in X$ having a label $y \in Y$.
The smaller PCS($x$,$y$) suggests $F$ has low confidence that $x$ has a label $y$.
\subsection{Explainability. }
Explainability refers to the capability of comprehending and interpreting how a machine learning model reaches its decisions or forecasts. 
Let $F$ be a machine learning model that takes an input $X$ and produces an output $y$. We can represent this relationship as $y=F(x)$
To assess the explainability of $F$, we need to examine how it makes its predictions. One way is to decompose $F$ into simpler and more easily understood functions. For example, we can use feature importance analysis to identify which input features have the strongest influence on the output.
Another approach is to use model-agnostic methods such as LIME (Local Interpretable Model-Agnostic Explanations) \cite{ribeiro2016shouldlime}  or SHAP (SHapley Additive exPlanations) \cite{lundberg2017unifiedSHAP} to generate local explanations for individual predictions. These methods identify which input features were most important for a particular prediction, and how changes to those features would affect the output.

%\garrison{note the space after bracket} 
\subsection{Pre-Trained Embedding. }
Pre-trained embeddings refer to using pre-existing word vectors that have been trained on large datasets using unsupervised machine learning techniques such as word2vec \cite{church2017word2vec}, GloVe \cite{pennington2014glove}, or fastText \cite{bojanowski2017fastext}. These embeddings are trained on large corpora of text data, and the resulting word vectors capture semantic and syntactic relationships between words. By using pre-trained embeddings, it is possible to leverage the knowledge captured in these embeddings to improve the performance of NLP tasks such as text classification, sentiment analysis, and machine translation without the need to train a new embedding from scratch.
\section{Related Work}
\label{relatedwork}
\begin{table}[bt!]
\begin{small}
\caption{Comparison of our EDIT with inference-time defenses}

\label{comparison}
\centering
\resizebox{\columnwidth}{!}{\begin{tabular}{|c|c|c|c|c|c|c|c|c|}
\hline
\textbf{Topic Covered} 
&\rotatebox{90}{{DISP\cite{zhou2019learning}}}
&\rotatebox{90}{{FGWS\cite{mozes2020frequency}}}
&\rotatebox{90}{{WDR\cite{mosca2022suspicious}}}
&\rotatebox{90}{{RDE\cite{yoo2022detection}}}
&\rotatebox{90}{{MLMD\cite{MLMTAED2023}}}
&\rotatebox{90}{{TextShield\cite{shen2023textshield}}}
&\rotatebox{90}{{SHAP \cite{huber2022detecting}}}
&\rotatebox{90}{\textbf{\underline{EDIT}}}\\
\hline 
Threshold-independent&\checkmark&$\times$&\checkmark&$\times$&\checkmark&\checkmark&\checkmark&\checkmark\\
\hline
Computationally Efficient Features&$\times$&\checkmark&$\times$&\checkmark&\checkmark&$\times$&$\times$&\checkmark\\
\hline

Use Explainability Features&$\times$&$\times$&$\times$&$\times$&$\times$&\checkmark&\checkmark&\checkmark\\
\hline
Offer Dynamic Correction&$\times$&$\times$&$\times$&$\times$&$\times$&$\times$&$\times$&\checkmark\\
\hline 
Provides Threat Intelligence&$\times$&$\times$&$\times$&$\times$&$\times$&$\times$&$\times$&\checkmark\\
\hline
\end{tabular}}
\end{small}
\end{table}
Our defence framework, EDIT, falls under the category of \textit{inference-time} defence. Therefore, in this section, we present an overview of SOTA inference defence methods and provide comparative analysis with EDIT. We summarize the comparison of inference-time defences  related to EDIT in Table~\ref{comparison}.

\textbf{Training-time defences.} These defences aim to establish resilient models capable of performing well on both clean and adversarial inputs \cite{li2021searching}. This category encompasses various defences, such as Adversarial Training \cite{wang2020infobert}, Synonym Encoding \cite{wang2021natural}, and Certifiable Robustness \cite{wang2021certified}. However, these approaches face significant challenges: (i) They often necessitate retraining TTC models by incorporating Adversarial Examples (AEs) into the training set, resulting in computational expense and time-consuming processes \cite{shen2023textshield}. (ii) Effectiveness may be confined to specific attacks, limiting their utility against sophisticated and unforeseen attack techniques \cite{li2021searching}. (iii) Some defences depend on external information, such as knowledge graphs, to enhance robustness \cite{le2022shield}. (iv) Unrealistic assumptions, like knowledge of the attacker's embedding method, are prevalent in many defences \cite{li2021searching}. These limitations underscore the formidable challenges in developing defences that truly enhance robustness in TTCs.
{Inference-time defences concentrate on identifying and obstructing AEs. For instance,  Yoo et al. \cite{yoo2022detection} utilize Robust Density Estimation (RDE) and anomaly detection for AE identification while Frequency-Guided Word Substitution (FGWS) \cite{mozes2020frequency} analyzes changes in word frequency distribution to pinpoint AEs. Another recent method, MLMD \cite{MLMTAED2023} utilizes a pre-trained masked language model to distinguish between normal examples and AEs by exploring changes in the manifold.
However, these methods \cite{mozes2020frequency,yoo2022detection,MLMTAED2023} use \textit{threshold-based} techniques. While these detectors are simple to implement and computationally efficient, their effectiveness depends on precise threshold setting and adaptability to diverse data patterns which makes them less suitable for complex, dynamic, or high-dimensional real-world scenario.
Only two prior approaches \cite{shen2023textshield,huber2022detecting} utilize explainabilty measures to distinguish AEs from benign examples
Huber et al. \cite{huber2022detecting} leverage Shapley additive explanations whereas TextShield \cite{shen2023textshield}, a recent study, employs diverse saliency measures to compute Adaptive Word Importance (AWI) for training an adversarial detector. 
However, the approaches \cite{huber2022detecting,shen2023textshield} are not specifically designed for the complex architecture of transformers. They lack the utilization of attention cues inherent in transformer architecture during the computation of saliency factors, rendering them computationally intensive as illustrated by Chefer et al. \cite{chefer2021transformer}.
Another related work, Word-level Differential Reaction (WDR) \cite{mosca2022suspicious} investigate logit variations to train an AE detector. 
Notably, more comprehensive defence mechanisms, DIScriminate Perturbation (DISP) \cite{zhou2019learning} and TextShield \cite{shen2023textshield} integrate detection and correction, but they offer static correction mechanisms. 
For example, TextShield replaces perturbed words with high-frequency synonyms, while DISP replaces them with the most contextually similar neighbors. % However, these methods lack interpretability and fail to provide security analysts with essential threat intelligence or identify scenarios where correction mechanisms may falter, hindering the identification of vulnerabilities in both the model and its defence mechanisms \cite{shen2023textshield,zhou2019learning}. \updated{Compared to TextShield~\cite{shen2023textshield}, our approach differs in both methodological and practical aspects. TextShield integrates LSTM-based components and post-hoc classifiers to detect adversarial sentences, while EDIT applies a lightweight statistical detector and a \textit{dynamic transformation module} that adaptively replaces perturbed words using context-aware masked language modeling. Furthermore, EDIT incorporates a human-in-the-loop alerting mechanism and threat intelligence extraction, enhancing transparency and robustness. These distinctions make EDIT not only faster and more interpretable but also operationally more holistic than TextShield.} 
% Therefore, \textit{none of the aforementioned work provide a holistic, interpretable and comprehensive inference-time defence against 
\updated{
\noindent \textbf{Comparison to TextShield.} While both \textsc{EDIT} and TextShield~\cite{shen2023textshield} use explainability for adversarial detection, \textsc{EDIT} introduces several key advancements. First, for adversarial detection, it leverages transformer-specific attention gradients for relevance estimation, enabling model-aligned, context-sensitive attribution—unlike TextShield’s use of model-agnostic SHAP scores. Importantly, \textsc{EDIT} employs a lightweight statistical detector (XGBoost) trained on extracted statistical features, which is significantly more efficient than the LSTM-based classifier used in TextShield.
Second, \textsc{EDIT} automates perturbed word identification via outlier detection over multiple features (e.g., attribution score, prediction shift, word rarity), in contrast to TextShield’s static thresholding.
Third, \textsc{EDIT}'s transformation phase dynamically selects replacements from semantic, lexical, and contextual candidates using PCS feedback, whereas TextShield performs static synonym substitution (e.g., using NLTK) without model feedback.
Finally, \textsc{EDIT} supports human-in-the-loop alerting, offering greater transparency and operational control.
These innovations collectively make \textsc{EDIT} a more robust, efficient, and interpretable defense system compared to prior methods.
}

We have compared EDIT with four SOTA defences including WDR \cite{mosca2022suspicious}, SHAP \cite{huber2022detecting}, RDE \cite{yoo2022detection} and MLMD \cite{MLMTAED2023}.

\begin{figure}[tb!]
\centerline{\includegraphics[width=\columnwidth]{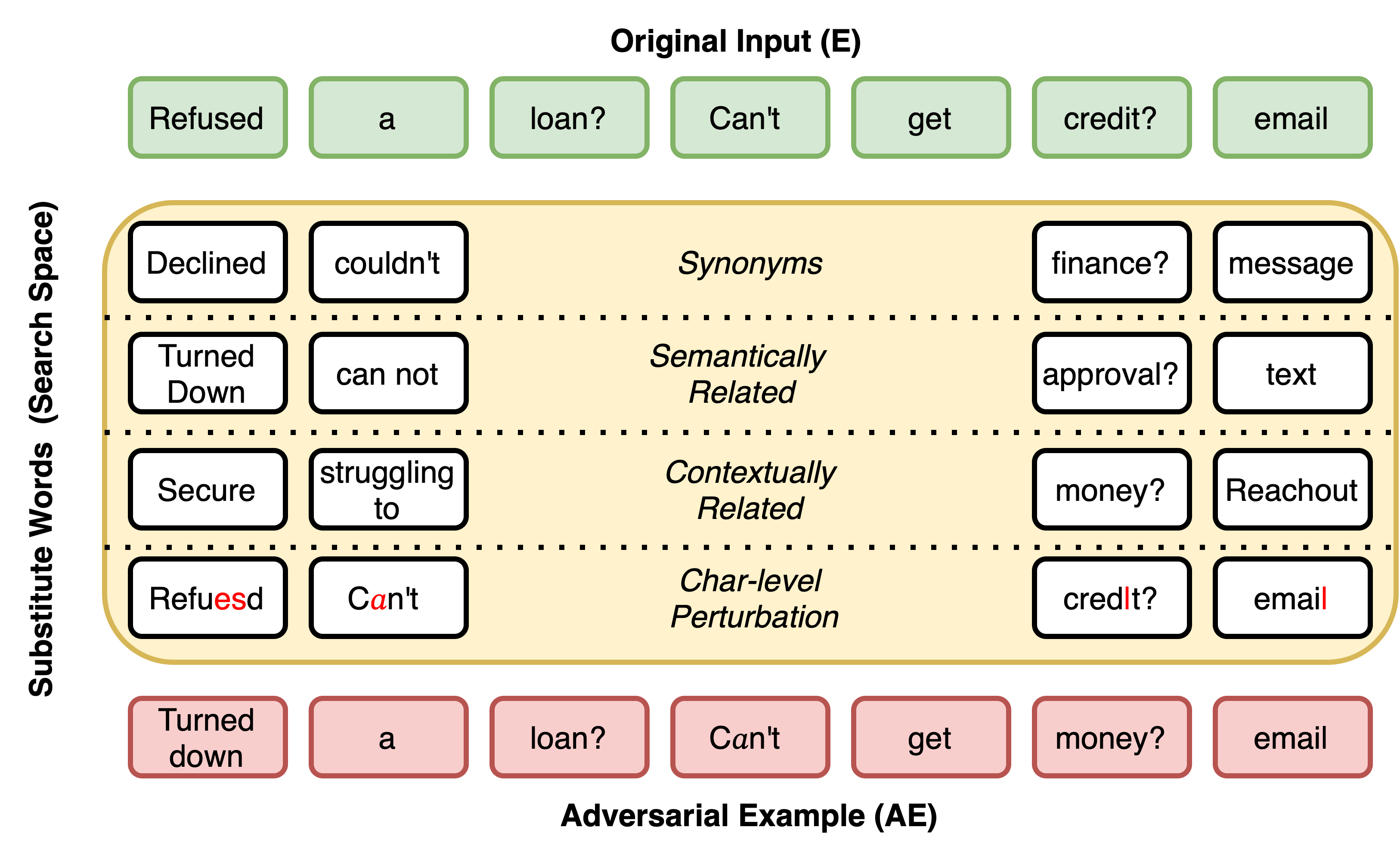}}
\caption{Word substitution attack model}
\label{ThreatModel}
\end{figure}
\section{Threat Model}
% \garrison{If we only consider WSA attack, we need to briefly say why. is this the most used AE attack method in textual AE? Better say `following xxxx works, we mainly focus on WSA attack...'}

\textbf{Attacker's Goal and Knowledge.} Building upon SOTA studies \cite{MLMTAED2023,shen2023textshield,yoo2022detection,mosca2022suspicious}, this paper adopts Word Substitution Attack (\texttt{WSA}) as the evasion threat model against Transformer-based Text Classifiers (TTC). As illustrated in Fig~\ref{ThreatModel}, in a WSA the attacker strategically replaces a subset of important words $W$ in the original input $E$ with $W'$ (e.g., synonyms, semantically or contextually similar or misspelled version) to craft an Adversarial Example (AE) with an objective is to fool $T_m$ into making incorrect predictions such that ($T_m(E) \neq T_m(AE)$) while preserving the original text's meaning \cite{li2018textbugger}.
Recent studies have shown that TTC can misclassify over 90\% of AEs generated using WSA, highlighting the potential impact on users who rely on the model's output \cite{li2020bert}.

{\textbf{Defender's Goal and Knowledge.} Consequently, we propose an inference-time defence called \texttt{EDIT} against WSA with a goal to provide holistic protection. Our approach includes detecting $AE$, identifying perturbed words ($P_\mathrm{cand}$), transforming $AE$ into benign examples ($TF_E$), and notifying analysts about unsuccessful conversions. Assuming complete knowledge of $T_m$ model internals, we develop a differentiator ($M_\mathrm{ad}$) for $AE$ and clean examples ($CE$), an identifier for perturbed words ($W'$), and a transformation and alerting module. }
% \garrison{This should be `Defender's Goal and Knowledge'}
\section{EDIT Framework}
\begin{figure}[tb!]
\centerline{\includegraphics[width=\columnwidth]{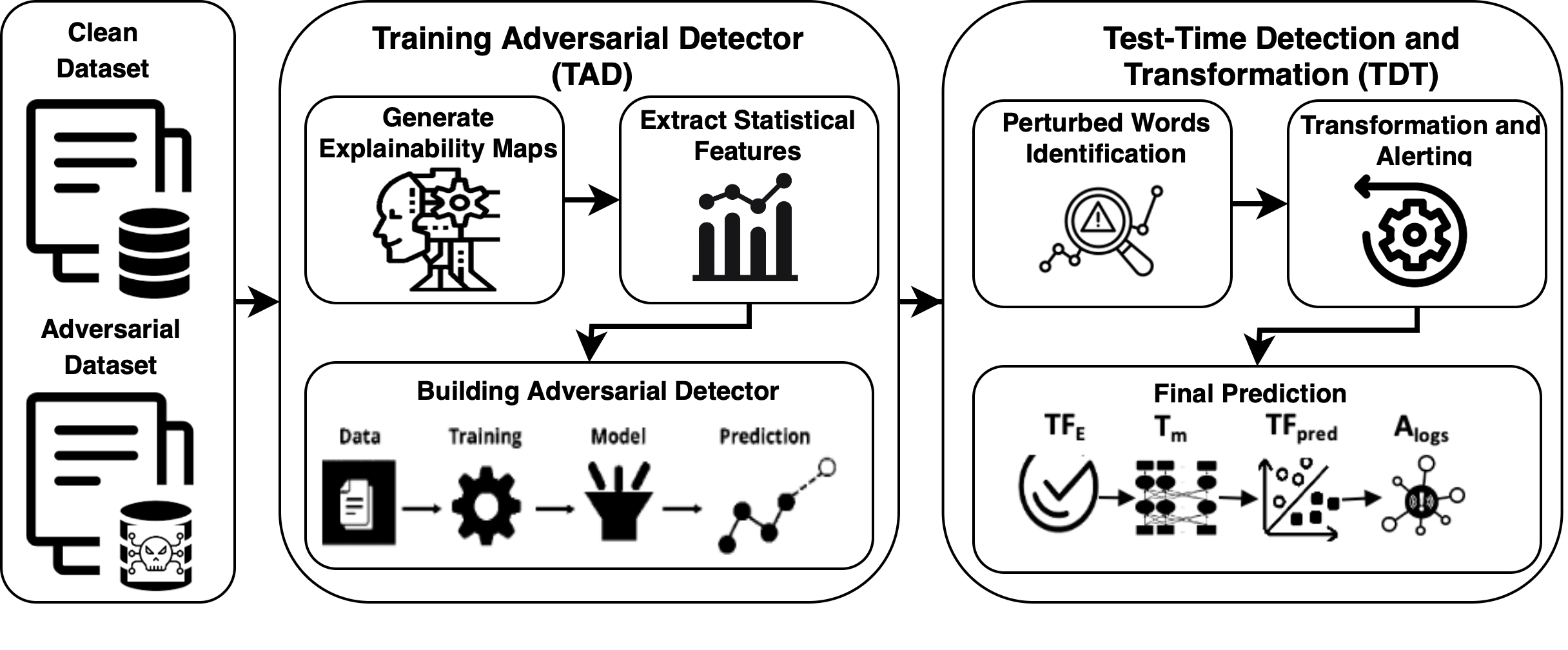}}
\caption{Overview of EDIT framework}
\label{figure_overview}
\end{figure}
Fig \ref{figure_overview} overviews the EDIT framework that consists of two main phases:
Training Adversarial Detector (TAD) and Test-Time Detection and AE Transformation (TDT). 

\subsection{Training Adversarial Detector (TAD).}
The TAD phase aims to construct a machine learning (ML) model that discriminates between CEs and AEs. 
To achieve this objective, TAD takes the original and adversarial datasets as inputs.
%Note that the training of adversarial detection can work without seeing any adversarial datasets (as validated in Section \ref{unknown_attacks}), but we focus on leveraging a small set of adversarial dataset for adversarial detector training to boost its performance even against AEs during test time crafted with a multitude of unseen AE creation methods.
The original and adversarial examples are inputs to generate explainability maps step that back-propagates and aggregates attention weights and gradients across all layers and heads of surrogate $T_m'$ model to generate attention maps ($A_\mathrm{map}$) and integrated gradient ($I_\mathrm{grad}$) distributions. The $T_m'$ enables the detection of AEs without impeding the regular functioning of the original model $T_m$. 
The $A\mathrm{map}$ and $I\mathrm{grad}$ are then sent to next step that extracts statistical features from them and additionally computes frequency-based features from the datasets to construct a feature set for training several supervised ML classifiers as Adversarial Detector ($M_\mathrm{ad}$) candidates. These classifiers undergo fine-tuning using K-fold cross-validation, and the most effective intermediate models are selected for each ML model type. Finally, an unseen test dataset is employed to evaluate the selected models, and the ML model exhibiting the highest validation performance is designated as the $M_\mathrm{ad}$.
\begin{algorithm}[tb!]
\caption{Generate explainability map}
\label{explain}
{\bfseries Input:} Pre-trained model $T_m'$, training dataset $T_{data}$ with $k$ examples;
{\bfseries Output:} ${A_{\mathrm{map}}}$, ${I_{\mathrm{grad}}}$;

Initialize ${A_{\mathrm{map}}}$ and ${I_{\mathrm{grad}}}$ to $\mathit{I}^\mathrm{nxn}$;\\
\ForAll{$E$ in $T_data$}{
$y_\mathrm{pred}$=$T_m(E)$\\
\ForAll{$l$ in $T_m.layers$}{
Compute attention weights using the inputs query $q$ and key $k$ vectors with $d$ embedding dimensions;\\
$A_\mathrm{w}^l \leftarrow \frac{\text{softmax}(q.k^T)}{\sqrt {d}}$\\
Compute gradient of attention with respect to $y_\mathrm{pred}$;\\
$\nabla A_\mathrm{w}^l=\frac{\partial y_\mathrm{pred} }{\partial A_\mathrm{w}^l}$\\
Compute weighted averaged cross all the heads in $l$ range [0,max], clamping negative contributions;\\
$A_\mathrm{map}=A_\mathrm{map}+\overline{(A_\mathrm{w}^l \cdot \nabla A_\mathrm{w}^l)}$\\
$I_\mathrm{grad}=I_\mathrm{grad}+\overline{(\nabla A_\mathrm{w}^l)}$\\
}}
\textbf{return} ${A_{\mathrm{map}}^{dxk}}$, ${I_{\mathrm{grad}}^{dxk}}$
\end{algorithm}
\subsubsection{Generate Explainability Maps. } 
\label{EFFE}
In this step, we calculate $A_\mathrm{maps}$ and $I_\mathrm{grad}$ distributions for each word $n$ in an input sequence $E$. $A_\mathrm{maps}$ gauges word importance based on attention, reflecting weights for a prediction $y_{pred}$. $I_\mathrm{grad}$ assesses input word significance, revealing sensitivity. Inspired by recent work in computer vision \cite{chefer2021generic}, our method efficiently computes these distributions without additional training or modification, extending to differentiate original from adversarial examples in self-attention-based transformer models popular in text classification.
Algorithm \ref{explain} summarizes the generation of explainability maps. Initially, $A_\mathrm{map}$ and $I_\mathrm{grad}$ are initialized with an identity matrix $I_\mathrm{n*n}$ to denote the self-importance of each word in the input sequence $E$. Attention weights $A_\mathrm{w}^l$ across each layer of the $T_m'$ model reflect the relative importance of each word or sub-word in $E$. These $A\mathrm{w}^l$ are computed using the dot product of a query vector $q$ and key vectors $k$, followed by softmax normalization. Attention gradients $\nabla A_\mathrm{w}^l$ are obtained by computing the derivative of the output $y_\mathrm{pred}$ with respect to the attention weights $A_\mathrm{w}^l$.
$A_\mathrm{maps}$ is determined by the Hadamard product of attention weights and gradients, averaged across all heads in $l$ after excluding negative contributions. Similarly, $I_\mathrm{grad}$ is calculated by averaging $\nabla A_\mathrm{w}^l$ across all heads in $l$, clamping negative values. The module outputs $d*k$ dimensional results, providing $A_\mathrm{map}$ and $I_\mathrm{grad}$ for all examples $k$.
\subsubsection{Extract Statistical Features. } 
We employ various statistical measures \cite{merlo2001automatic}, including minimum, maximum, mean, median, mode, variance, skewness, kurtosis, entropy, mean of gradient, and sum of peaks, to extract information on central tendency, variability, and distribution shape from four types of distributions for a given text sample $E$. These distributions include $A_\mathrm{map}$, representing overall attention received by words in $E$; the $I_{grad}$ distribution, comparing the relative importance of words across different predicted labels $y_{pred}$ and capturing differences between expected and adversarial samples; the Overall frequency distribution, characterizing word frequency in $E$ relative to the training dataset; and the outlier frequency (OF) distribution, indicating the frequency of outliers in $A_\mathrm{map}$. Outliers are identified using the inter-quartile range (IQR) method, and their frequency distribution is obtained, providing insights into the frequency of outliers in the attention map \cite{mozes2020frequency}.
% (details about the feature importance in provided in Fig\ref{FeaturesImp} in appendix)

\subsubsection{Building Adversarial Detector. } To build the adversarial detector, firstly, we  pre-processed the feature set ($F_\mathrm{set}$) to address any inconsistencies and redundancies, thus ensuring data integrity. Then, we leverage a diverse set of machine learning models and optimized them based on metrics especially designed for imbalance datasets. We then identify the optimal AE detection model ($M_\mathrm{ad}$) through 10-fold cross-validation. This selection process is designed to prevent over-fitting and is followed by a thorough evaluation of $M_\mathrm{ad}$ on an unseen dataset to validate its efficacy in distinguishing AEs from CEs. This comprehensive approach underscores the robustness and reliability of the AE detection system \cite{sabir2022reliability}.

\subsection{Test-Time Detection and AE Transformation (TDT). } The TDT phase aims to perform four main task: (i) differentiate $CE$ from AE using $M_\mathrm{ad}$, (ii) identify perturbed words in the detected $AE$, (iii) transform $AE$ into an optimal non-adversarial variant ($TF_E$) preserving semantic similarity, and (iv) generate threat intelligence logs, alerting analysts if transformation fails. At test time, data is sent to $M_\mathrm{ad}$ for AE detection. Detected $AE$ examples undergo perturbed word identification, generating candidate perturbed words ($P_\mathrm{cand}$). These candidates are transformed by selecting optimal substitutions from a set of $N$ replacement options ($S_\mathrm{cand}$) using pre-trained embeddings. The reverse-engineered $AE$, $TF_E$, and $CE$ undergo spelling-based correction. If $TF_E$ is still predicted as $AE$ and $T\_m'(AE) = T_m'(TF_E)$, an alert is created, providing threat intelligence logs ($A_\mathrm{log}$). $TF_E$ is then sent to the original $T_m$ model for final prediction.

\subsubsection{Perturbed Words Identification. }
This step identifies potentially perturbed words in adversarial examples (AEs) using a multifaceted feature-based approach. It includes explainability scores, model-based metrics, word frequency, grammatical analysis, logistic regression-derived importance scores, and label change detection. The final phase involves training, tuning, and evaluating multiple classifiers to detect perturbed words ($P_\mathrm{Cand}$) in AEs, selecting the optimal classifier based on performance metrics. This automated model streamlines detection, providing a robust system surpassing state-of-the-art methods.

\textbf{Feature Extraction and Training. } 
We extract a set of token-level features designed to capture semantic, statistical, and syntactic signals indicative of adversarial perturbations:
\updated{
 \textbf{(i) Explainability Scores:} For each word in the AE and its adjacent context (i.e., immediate neighbors), explainability scores ($A_\mathrm{map}$) are computed. These scores measure the contribution of each token to the model’s decision, helping to highlight influential words. \textbf{(ii) Replacement Score ($\Delta PCS$):} We calculate the change in Prediction Confidence Score (PCS) when a word is masked. This quantifies how much the model’s prediction confidence drops when that word is removed—larger drops indicate greater influence.\textbf{(iii) Word Frequency:} We assess how frequently each word occurs in the clean training corpus. Rare words are more likely to be adversarial insertions and thus indicative of perturbation.\textbf{(iv) Grammatical and Syntax Errors:} This feature flags words that create grammatical inconsistencies or syntax violations, under the hypothesis that AEs often introduce such errors. \textbf{ (v) Logistic Regression Importance:} A logistic regression model is trained on clean data using TF-IDF vectors. Feature importance scores from this model serve as an auxiliary signal to identify unusually influential words. \textbf{(vi) Label Change Detection:} For each word, we observe whether its removal leads to a change in predicted label. Such words are marked as critical contributors to the adversarial behavior.
The extracted features are combined to form a rich representation of each word. All scalar features are normalized to a [0, 1] range. We then train multiple classifiers (e.g., Random Forest, XGBoost, Logistic Regression) to distinguish perturbed from non-perturbed tokens, using SMOTE to handle class imbalance. Cross-validation with stratified sampling ensures robustness and generalization. The best-performing classifier is selected based on balanced accuracy (BAL\_ACC) and F1-Score.}

\updated{ 
To detect perturbed words, we combine explainability scores (ExplainScore) and frequency features (FreqScore) in a multi-step process. The explainability score evaluates how much influence a word has on the model’s decision, using techniques like attention mechanisms or feature importance. On the other hand, the frequency score measures how unusual a word’s occurrence is in the adversarial dataset compared to the clean dataset.
To integrate these two features, we first normalize both explainability and frequency scores to a [0, 1] range, ensuring they are on a comparable scale. Then, we combine these scores using a weighted average approach, where each score contributes equally. This combined score allows us to capture both the semantic relevance (from ExplainScore) and statistical rarity (from FreqScore) of words.  
The resulting combined score is thresholded: if the combined score of a word exceeds a predefined threshold, that word is flagged as perturbed. This approach ensures that words exhibiting both high importance to model decisions and abnormal frequency distributions are effectively detected as adversarial perturbations.}

\begin{algorithm}[!tb]
\small
\caption{Transformation and alerting}
\label{algo2}
\SetAlgoLined
\KwIn{$AE$, $P_{cand}$, $S_{cand}$, $PCS[Y']$, $Y'$, $adv_{flag}$, $adv_{org}$}
\KwOut{$TF_E$, $Human$, $HumanMsg$}

\For{$try$ in $Trials$}{
    \For{$cand$ in $P_{cand}$}{
        $TF_E \leftarrow AE$\\
        Generate transformations of $AE$ by replacing $cand$ with $S_{cand}$\\
        $r\_PCS, r\_Y' = T_m(TF_{cand})$\\
        $r_{score} = PCS[Y'] - r\_PCS[Y']$ \\
        $freq_{score} \leftarrow \text{frequency}(S_{cand})$\\
        $sub\_simi\_score = \text{cosine\_similarity}(AE, TF_E)$
        
        $opt_{score} = r_{score} + sub\_simi\_score + simi_{score} + freq_{score}$\\
        \textbf{$sel_{id} = \arg\max(opt_{score})$\\}

        \If{$r_{score}[sel_{id}] > 0$ \textbf{or} $r\_Y'[sel_{id}] \neq Y'$}
        {
            Select optimal $TF_E$ and $cand_{opt}$\\
            $adv\_{Y'}, adv_{score} = M_{ad}(TF_E)$\\
            \If{$adv_{Y'} = 0$ \textbf{and} $r\_Y'[sel_{id}] \neq Y'$}{
                $adv_{flag} \leftarrow \text{False}$, $Human \leftarrow \text{False}$, $HumanMsg \leftarrow \text{Converted to non-Adv}$\\
                \KwRet $TF_E$, $Human$, $HumanMsg$\\
            }
        }
    }

    \If{$adv_{score} > adv_{org}$}{
        $Human \leftarrow \text{True}$, $HumanMsg \leftarrow \text{Got more Adv}$\\
        \KwRet $TF_E$, $Human$, $HumanMsg$\\
    }

    \If{no replacement}{
        $Human \leftarrow \text{True}$, $HumanMsg \leftarrow \text{No optimal replacement found}$\\
        \KwRet $TF_E$, $Human$, $HumanMsg$\\
    }
}
\If{$adv_{flag} = \text{True}$}{
    $Human \leftarrow \text{True}$, $HumanMsg \leftarrow \text{Couldn't convert to non-Adv}$\\
    \KwRet $AE$, $Human$, $HumanMsg$\\
}
\end{algorithm}

 \subsubsection{Transformation and Alerting. }
The algorithm takes as input the detected adversarial example (AE) and identified perturbed words from the identifier module ($P_\mathrm{cand}$), sorted by the confidence of the identifier module on the word being perturbed.

\emph{Transformation. }
     This module is responsible for neutralizing the adversarial effects of AE. For each word in $P_\mathrm{Cand}{i}$, the algorithm explores potential substitutions ($S_\mathrm{Cand}$) using three diverse embedding techniques (synonyms, semantic, and contextual) to ensure dynamic and varied substitution options.
     The algorithm computes the candidate ($cand_\mathrm{opt}$) with the maximum optimization score ($opt_\mathrm{score}$) using the equation:
     \begin{equation}
       opt_\mathrm{score} = \Delta PCS(\text{Y'}) + sub\_simi_\mathrm{score}+ simi_\mathrm{score} + freq_\mathrm{score}.
     \end{equation}
    
    Here $\Delta PCS(\text{Y'})$ represents the change in the PCS when $w_i$ in $P_\mathrm{Cand}$ is replaced by $s_k$ in $S_\mathrm{Cand}$. $sub\_simi_\mathrm{score}$ denotes the semantic similarity between the detected AE and $TF_E$ formed after substitution.
     $simi_\mathrm{score}$ reflects the similarity between $w_i$ and substitution $s_i$.
     $freq_\mathrm{score}$ indicates the frequency of $s_k$ in the original training dataset.
     If $r_{score}[sel_{id}] > 0$, it means that the optimal substitution reduces the confidence on the current label or the label is changed, indicating a suitable $TF_E$ for the next substitution.
     If the selected example ($TF_E$) is less adversarial than the original ($AE$) based on the adversarial detector $M_\mathrm{ad}$, it replaces $AE$ with the chosen example. If $M_\mathrm{ad} (TF_E)= CE$ and $T_m'(AE) \neq T_m'(TF_E)$, then the transformation completes.
     
 \emph{Alert Generation for Security Analyst Intervention. } This sub-module effectively engages human analysts when automated processes fail, adding a layer of reliability and reducing the risk of mis-classifications going unnoticed. An alert is generated for the security analyst if, after all trials and $P_\mathrm{cand}$ replacements, the detector still classifies the example as adversarial or if the label of the transformed example remains the same as the originally classified adversarial example. The module also provides threat intelligence logs $A_\mathrm{log}$ and an alert message to the analyst to facilitate prompt and informed intervention.

 \emph{Spelling Correction. } Finally, both $TF_E$ and $CE$ undergo spelling correction to rectify any typos for non-pronoun words not present in the training dataset of $T_m'$.

\section{Experimental Setup}
\label{chapter6_Results}
In this section, we describe the comprehensive setup of our experiments, the datasets used and the performance metrics adopted to evaluate our approach. We conducted our experiments on the Google Colab Pro+ version, 16 GB RAM and T4 GPU. To ensure research reproducibility, we made our codebase available to the research community \footnote{\url{https://shorturl.at/bvxz4}}.

\subsection{Datasets and Victim Models. } 

% \textbf{Transformer Models and Baseline Datasets}.
% We selected two prominent Transformer-based architectures, 
% (i) BERT \cite{devlin2018bert} and (ii) ROBERTA \cite{liu2019roberta}. These models are widely used and have demonstrated high accuracy in text classification \cite{wolf2020transformers}. While other transformer-based Large Language Models (LLMs) can be repurposed for classification, specialized models like BERT or RoBERTa are explicitly tailored and pre-trained for tasks involving text classification, often yielding superior performance \cite{bosley2023we}. They have also been adapted in the cyber-security domain for detecting various attacks, such as phishing and spam \cite{bayer2022cysecBERT,akbar2022knowledge,vladescu2021latest,lee2021d}. Due to their vulnerability to attacks and high attack success rates \cite{azizi2021t,zhou2020detection,chen2021badnl}, they are suitable for evaluating EDIT.
% These models are trained over \texttt{four} baseline datasets: IMDB, AGNEWS, YELP, and SST2. These are the most popular datasets used by SOTA  \cite{MLMTAED2023, yoo2022detection,mosca2022suspicious} to evaluate their work. The pre-trained models for BERT and ROBERTA corresponding to each of these datasets were obtained from \footnote{\url{https://huggingface.co/textattack}} to serve as the target Transformer-based Text Classification (TTC) models in our study. Table~\ref{Chapter6_dataset} for more details on the datasets. The datasets used in our experiments cover diverse topics and lengths to ensure generalizability.

We selected the Transformer architectures BERT \cite{devlin2018bert} and ROBERTA \cite{liu2019roberta} for their superior accuracy in text classification, outperforming other Large Language Models (LLMs) \cite{wolf2020transformers, bosley2023we}. Widely adopted in the cybersecurity domain, these models have been effective in detecting attacks such as phishing and spam \cite{bayer2022cysecBERT, akbar2022knowledge, vladescu2021latest, lee2021d}. Vulnerable to attacks and with high attack success rates \cite{azizi2021t, zhou2020detection, chen2021badnl}, they are suitable for evaluating EDIT. Trained on four baseline datasets (IMDB, AGNEWS, YELP, and SST2), these models, obtained from \footnote{\url{https://huggingface.co/textattack}}, serve as target Transformer-based Text Classification (TTC) models in our study. These datasets are widely used in SOTA research \cite{MLMTAED2023, yoo2022detection, mosca2022suspicious} and they cover diverse topics and lengths to ensure generalizability.

\subsection{Adversarial Attacks. }
\label{Attacks_Studied} 
This study explores adversarial attacks across three granularity levels: char-level, word-level, and multi-level. We employ the TextAttack Library \cite{morris2020textattack} and utilize an adversarial dataset from \cite{yoo2022detection}. The following attacks are employed. 
(i) Char-level attack: DeepWordBug (DWB) \cite{wang2018deep} utilizes a deep recurrent neural network to generate character-level perturbations.
(ii) Word-level attack Four word-level attacks are employed: (a) TextFooler (TF) \cite{jin2019bert}, (b) Bert\_Attack (BAE) \cite{garg2020bae}, (c) Probability Word Saliency Substitution (PWWS) \cite{ren2019generating}, and (d) TextFooler Adjusted (TF\_adj) \cite{yoo2022detection}. (e) Attack to Training: A2T uses gradient-based search to identify essential words in the text. It replaces them with similar words based on word embeddings while preserving Part of Speech (POS) and semantic similarity using Distil-Bert similarity scores (we tested it only in TDT stage as novel attack).
(iii) Multi-level attack: TextBugger (TB) \cite{li2018textbugger} generates adversarial examples by modifying text at the character, word, and phrase levels.

\emph{Seen Attacks. }
We used AEs from three attacks: BAE, DWB, and PWWS for training both the detectors and identifier. These attacks were selected for their diverse tactics: BAE  \cite{garg2020bae} manipulates contextual embeddings, DWB targets character-level changes, and PWWS implements synonym substitutions based on WordNet \cite{miller1995wordnet}. 

\emph{Unseen Attacks. }
For testing, we used samples from the previously mentioned seen attacks (BAE, DWB, and PWWS) and introduced \emph{unseen} novel attacks, including TextFooler, TextBugger, and tf-adj. This was done to evaluate the robustness and generalizability of the trained models. 
A2T was only used in TDT step to check the robustness of our method while testing. \new{\textbf{This setup is consistently applied for training and evaluating both the adversarial detector and the perturbed word identifier, ensuring a fair comparison of their generalization performance.}}

\subsection{TAD Setup. }
\label{TAD_SETUP}
The experimental setup used for the training adversarial detector is described below:

\texttt{(i) Training}. We used 1000 CEs and 1000 AEs, with each training attack type having around 1000/3 samples. 
Then we merged both clean and AEs and performed a stratified split based on attack vectors, maintaining an 80:20 training-test ratio. 
\texttt{(ii) Hyper-parameter tuning and classifiers}.
We trained five effective classifiers: XGBoost, LightGBM, RandomForest, Decision Tree, and Logistic Regression—as described in \cite{bonaccorso2018machine}. 
To select optimal traditional ML models, we applied Bayesian optimization \cite{snoek2012practical} using hyperopt library \cite{bergstra2013making}.
We chose bayesian optimisation because it is robust to noisy objective function evaluations \cite{wang2013bayesian}.
\texttt{(iii) Comparison with SOTA}.
We compared our approach with four SOTA methods SHAP \cite{huber2022detecting}, RDE \cite{yoo2022detection}, WDR \cite{mosca2022suspicious} and MLMDM \cite{MLMTAED2023} methods. 
For fair comparison, we maintained a same experimental setup for WDR, SHAP, and our EDIT method. However, for the RDE method, which relies on anomaly detection, we trained the detector on 80\% of the clean dataset and tested it on 100\% of the AEs along with 20\% CEs.
\texttt{(iv) Performance Metrics}.
We utilised the average Matthew Correction Coefficient (MCC) of 10-fold cross-validation with stratified sampling \cite{pmlr-v48-liberty16} (that retains a balanced class distribution in each fold) and early stopping criteria to select the optimal parameters.
% if within 100 trials there is no improvement in MCC. 
MCC was used to select the optimal model because it accurately evaluates binary classification performance in the presence of class imbalance by considering all aspects of the confusion matrix. \cite{luque2019impact} and is resilient against the
imbalance datasets \cite{luque2019impact,chicco2020advantages,bergstra2013making}. 
MCC value ranges from -1 to 1, a value close to -1, 0, 1 depicts a poor model (misclassify both classes), random model (classify both classes randomly) and a good model (classify both classes well), respectively.
Furthermore, to report the performance of the $M_{adv}$, we have used MCC, accuracy (ACC), balanced accuracy (BAL\_ACC), AUC, F1-Score, Precision, Recall, FPR and FNR.
Previous studies \cite{luque2019impact,chicco2020advantages,bergstra2013making} have shown that these metrics summarise the overall performance of ML models better than other measures especially .
We also used Feature Extraction Time (FET) to quantify the computation efficacy.
\subsection{TDT Setup. }
This section details the experimental setup used for the TDT module.

\subsubsection{Perturbed Words Identification. } 
\texttt{(i) Training.}
We used a dataset consisting of 1,000 adversarial examples (AEs), with each type of attack contributing approximately 333 samples. To address the significant imbalance between clean and perturbed words, we employed SMOTE, an up-sampling technique that generates synthetic examples to enhance the representation of the minority class and improve model generalization \cite{fernandez2018smote}. \new{Consistent with the adversarial detector setup (Section \ref{TAD_SETUP}), the identifier was trained on AEs from three \emph{seen} attacks (BAE, DWB, and PWWS) and evaluated on both seen and \emph{unseen} attacks (e.g., TF, TB, TF-ADJ). This setup allows us to explicitly measure the generalization capability of the identifier across perturbation strategies.  }

\texttt{(ii) Comparison with SOTA.}
Our method was compared against three state-of-the-art techniques for detecting perturbed words:  
(a) Replacement Score (ReplaceScore): As used in DISP by Zhou et al. \cite{zhou2019learning}, this method evaluates the effect of word masking on model confidence about the label.  
(b) Frequency (FreqScore): Implemented in FGWS by Mozes et al. \cite{mozes2020frequency}, this approach leverages word frequency analysis to identify potential manipulations.  
(c) Explainability Score (ExplainScore): Applied in TextShield by Shen et al. \cite{shen2023textshield}, this technique quantifies the influence of each word on the model’s decision, highlighting vulnerable words.  

\texttt{(iii) Performance Metrics.}
We report average balanced accuracy (BAL\_ACC) obtained via 10-fold cross-validation with stratified sampling \cite{pmlr-v48-liberty16}, ensuring balanced class distribution in each fold. Early stopping was used to optimize model parameters. BAL\_ACC was selected as the primary metric, as it provides a fair measure across classes and reflects true model performance irrespective of class imbalance \cite{tarzanagh2023fairness}. Additional metrics included ACC, AUC, F1-Score, Precision, Recall, FPR, and FNR, which are widely used in machine learning model evaluation \cite{luque2019impact, chicco2020advantages, bergstra2013making}.

\subsubsection{Transformation. }

To assess the TDT module's effectiveness, we randomly selected 200 AEs from each of the seven attacks detailed in Section \ref{Attacks_Studied}, and 200 clean examples (CEs) from the testing data. These examples were processed through the TDT module for evaluation. 
For optimal word replacement, our Dynamic Word Replacement algorithm \ref{algo2} utilizes:
(a) WordNet Synsets \cite{miller1995wordnet} for synonym-based replacements.
    (b)  GloVe Embeddings \cite{pennington2014glove}, specifically the "Glove-wiki-gigaword-50" trained on Wikipedia 2014 and Gigaword 5 data.
   (c) BERT’s Mask Language Model to identify contextually relevant replacement options.

For each method, we select the top 10 candidates to balance computational complexity and effectiveness.
\texttt{(i) Performance of Transformation. }
To evaluate the performance of the transformation module, we utilize the accuracy of transformation \(Acc_{transform}\) (equation~\ref{transformacc}), which measures how effectively the transformation module converts adversarial examples (AEs) to non-adversarial examples (non-AEs). Here, the notation \(N_{c}\) refers to the number of examples that were transformed from their adversarial label to their true label, and \(N_{det}\) is the number of examples detected as adversarial by the detector module.
\begin{equation}
\label{transformacc}
Acc_{transform} = \frac{N_{c}}{N_{det}}
\end{equation}
\texttt{(ii) Performance of Alerting. }
During the TDT phase, logs including example ID, \(M_\mathrm{ad}\) score, \(P_{cand}\), transformed text, ground truth label, \(y_{pred}\), \(TF_{pred}'\), confidence levels, human intervention flags, messages, and adversarial flags are collected. To alert security analysts, various messages are generated based on the outcome, ranging from requiring human intervention to successful or failed transformations. The efficiency of these alerts is quantified using \(Acc_{alert}\) (equation~\ref{humanacc}). Here, \(HI\) is a Boolean variable that indicates whether there is an alert or not, \(N_{in}\) and \(N_{c}\) represent the number of incorrectly and correctly classified examples respectively, and \(N_{det}\) is the number of examples detected as adversarial by the detector module.
\begin{equation}
\label{humanacc}
Acc_{alert} = \frac{(N_{det} \land N_{c} \land \neg HI) + (N_{det} \land N_{in} \land HI)}{N_{det}}
\end{equation}

\subsubsection{Overall Performance of TDT module. }
We comprehensively assess both the efficacy of automated transformations and the accuracy of final label predictions, offering a comprehensive overview of the system's performance. We calculate the overall accuracy without and with the alert module using the following equations, respectively. Additionally, we determine the total computational time taken by the TDT module, encompassing the detection to transformation and final prediction sub-stages.

\begin{equation}
\label{overallaccuracywohumanTDT}
Acc_{\mathrm{without\_alert}} = \frac{\sum_{i=1}^{n} (GT_i = TF_{i}) }{n}
\end{equation}

\begin{equation}
\label{overallaccuracyTDT}
Acc_{\mathrm{overall}} = \frac{\sum_{i=1}^{n} \left[(GT_{i} \neq TF_{i} \land H_i) \lor (GT_i = TF_{i}) \right]}{n}
\end{equation}

\section{Evaluation Results}
\label{sec:eval}
 The results of our investigation are presented in the subsequent subsections.

\begin{table*}
\centering
\begin{small}
\caption{Performance comparison with SOTA adversarial example detection methods (\updated{FET stands for feature extraction time)}}
\label{Performance_Comparison_Small_Scale}
\resizebox{\textwidth}{!}{\begin{tabular}{c|c|c|c|ccccccccc}
\hline
\textbf{Dataset} & \textbf{T\_m} & \textbf{Method} & FET (sec)& \textbf{MCC} & \textbf{ACC} & \textbf{BAL\_ACC} & \textbf{PRECISION} & \textbf{RECALL} & \textbf{F1-SCORE} & \textbf{AUC} & \textbf{FPR} & \textbf{FNR} \\
\hline
\multirow{10}{*}{\textbf{AGNEWS}}&\multirow{5}{*}{\textbf{BERT}}&\textbf{EDIT}&\textbf{0.07}&\textbf{0.77}&\textbf{88.28}\%&\textbf{88.27}\%&\textbf{88.36}\%&88.27\%&\textbf{88.27}\%&\textbf{88.27}\%&\textbf{14.00}\%&\textbf{9.45}\%\\
&&WDR&5.03&0.76&87.78\%&87.78\%&86.54\%&\textbf{89.55}\%&88.02\%&87.78\%&14.00\%&10.45\%\\
&&SHAP&14.4&0.57&78.05\%&78.07\%&81.92\%&72.14\%&76.72\%&78.07\%&16.00\%&27.86\%\\
&&RDE&0.09&0.42&69.80\%&77.88\%&97.05\%&65.77\%&78.41\%&77.88\%&10.00\%&34.23\%\\
&&MLMDM&1.25&0.34&64.28\%&64.23\%&80.43\%&	37.55\%&51.20\%&64.23\%&9.10\%&62.45\%\\
\cline{2-13}
&\multirow{5}{*}{\textbf{ROBERTA}}&\textbf{EDIT}&\textbf{0.06}&\textbf{0.72}&\textbf{85.79}\%&\textbf{85.78}\%&\textbf{85.86}\%&\textbf{85.78}\%&\textbf{85.78}\%&\textbf{85.78}\%&\textbf{16.50}\%&\textbf{11.94}\%\\
&&WDR&5.17&0.67&83.54\%&83.54\%&82.30\%&85.57\%&83.90\%&83.54\%&18.50\%&14.43\%\\
&&SHAP&15.94&0.49&74.1\%&74.1\%&81.3\%&62.7\%&70.8\%&74.1\%&14.5\%&37.3\%\\
&&RDE&0.08&0.38&66.14\%&75.69\%&96.85\%&61.38\%&75.14\%&75.69\%&10.00\%&38.62\%\\
&&MLMDM&1.15&0.32&63.13\%&63.07\%&80.66\%	&34.34\%&48.17\%&63.07\%&8.20\%&65.66\%\\
\hline
\multirow{8}{*}{\textbf{IMDB}}&\multirow{5}{*}{\textbf{BERT}}&\textbf{EDIT}&\textbf{0.15}&\textbf{0.82}&\textbf{90.77}\%&\textbf{90.76}\%&\textbf{91.01}\%&\textbf{90.76}\%&\textbf{90.76}\%&\textbf{90.76}\%&\textbf{13.00}\%&\textbf{5.47}\%
\\
&&WDR&8.43&0.76&87.78\%&87.77\%&85.85\%&90.55\%&88.14\%&87.77\%&15.00\%&9.45\%
\\
&&RDE&0.4&0.59&83.78\%&86.47\%&97.75\%&82.44\%&89.44\%&86.47\%&9.50\%&17.56\%
\\ &&SHAP&25.75&0.46&73.07\%&73.06\%&71.63\%&76.62\%&74.04\%&73.06\%&30.50\%&23.38\%
\\
&&MLMDM&26.08&0.35&65.83\%&65.79\%&77.54\%	&44.38\%&56.45\%&65.79\%&12.80\%&55.62\%\\
\cline{2-13}
&\multirow{5}{*}{\textbf{ROBERTA}}&\textbf{WDR}&7.92&\textbf{0.80}&\textbf{89.97}\%&\textbf{89.99}\%&\textbf{86.64}\%&\textbf{94.47}\%&\textbf{90.38}\%&\textbf{89.99}\%&\textbf{14.50}\%&\textbf{5.53}\%\\
&&EDIT&\textbf{0.11}&0.75&87.47\%&87.48\%&87.84\%&87.48\%&87.44\%&87.48\%&17.50\%&7.54\%\\
&&RDE&0.13&0.70&89.52\%&89.91\%&97.90\%&89.33\%&93.42\%&89.91\%&9.50\%&10.67\%\\
&&MLMDM&22.96&0.45&71.04\%&71.01\%&83.28\%&52.51\%&64.41\%&71.01\%&10.50\%&47.49\%\\
&&SHAP&23.23&0.28&64.16\%&64.15\%&65.38\%&59.80\%&62.47\%&64.15\%&31.50\%&40.20\%\\
\hline
\multirow{8}{*}{\textbf{SST2}}&\multirow{5}{*}{\textbf{BERT}}&\textbf{EDIT}&\textbf{0.06}&\textbf{0.78}&\textbf{89.06}\%&\textbf{89.01}\%&\textbf{89.16}\%&\textbf{89.01}\%&\textbf{89.04}\%&\textbf{89.01}\%&\textbf{13.47}\%&\textbf{8.50}\%
\\ &&WDR&3.95&0.65&82.29\%&82.29\%&81.25\%&84.08\%&82.64\%&82.29\%&19.50\%&15.92\%\\
&&SHAP&9.95&0.57&78.30\%&78.28\%&74.57\%&86.07\%&79.91\%&78.28\%&29.50\%&13.93\%\\
&&MLMDM&1.11&0.38&68.54\%&68.55\%&65.94\%&	76.41\%&70.79\%&68.55\%&39.30\%&23.59\%\\
&&RDE&0.07&-0.24&15.47\%&46.10\%&11.11\%&0.20\%&0.39\%&46.10\%&8.00\%&99.80\%\\
\cline{2-13}
&\multirow{5}{*}{\textbf{ROBERTA}}&\textbf{EDIT}&\textbf{0.05}&\textbf{0.77}&\textbf{88.49}\%&\textbf{88.46}\%&\textbf{88.55}\%&\textbf{88.46}\%&\textbf{88.48}\%&\textbf{88.46}\%&\textbf{13.54}\%&\textbf{9.55}\%\\
&&WDR&4.02&0.63&81.50\%&81.50\%&79.44\%&85.00\%&82.13\%&81.50\%&22.00\%&15.00\%\\
&&SHAP&11.84&0.41&70.3\%&70.3\%&68.8\%&74.0\%&71.3\%&70.3\%&33.5\%&26.0\%
\\ 
&&MLMDM&0.34&0.03&50.55\%&50.46\%&60.47\%&	2.61\%&5.00\%&50.46\%&1.70\%&97.39\%\\
&&RDE&0.07&-0.19&15.89\%&47.55\%&9.09\%&0.10\%&0.20\%&47.55\%&5.00\%&99.90\%\\
\hline
\multirow{8}{*}{\textbf{YELP}}&\multirow{4}{*}{\textbf{BERT}}&\textbf{EDIT}&\textbf{0.10}&\textbf{0.87}&\textbf{93.25}\%&\textbf{93.25}\%&\textbf{93.28}\%&\textbf{93.25}\%&\textbf{93.25}\%&\textbf{93.25}\%&\textbf{8.00}\%&\textbf{5.50}\%
\\
&&WDR&6.95&0.84&92.00\%&92.00\%&92.00\%&92.00\%&92.00\%&92.00\%&8.00\%&8.00\%
\\
&&RDE&0.13&0.44&72.12\%&79.27\%&97.16\%&68.54\%&80.38\%&79.27\%&10.00\%&31.46\%\\
&&MLMDM&11.95&0.40&68.36\%&68.28\%&81.21\%&	47.48\%&59.92\%&68.28\%&10.91\%&52.52\%\\
&&SHAP&19.01&0.35&67.25\%&67.25\%&66.51\%&69.50\%&67.97\%&67.25\%&35.00\%&30.50\%\\
\cline{2-13}
&\multirow{4}{*}{\textbf{ROBERTA}}&\textbf{EDIT}&\textbf{0.08}&\textbf{0.89}&\textbf{94.50}\%&\textbf{94.63}\%&\textbf{94.53}\%&\textbf{94.63}\%&\textbf{94.50}\%&\textbf{94.63}\%&\textbf{8.00}\%&\textbf{2.75}\%\\
&&WDR&6.99&0.82&90.84\%&90.90\%&88.89\%&92.31\%&90.57\%&90.90\%&10.50\%&7.69\%\\
&&MLMDM&10.98&0.40&67.87\%&67.84\%&82.69\%&	45.08\%&58.35\%&67.84\%&9.41\%&54.92\%\\
&&SHAP&20.32&0.40&70.16\%&69.67\%&72.97\%&59.34\%&65.45\%&69.67\%&20.00\%&40.66\%\\
&&RDE&0.13&0.26&51.42\%&66.84\%&95.61\%&43.69\%&59.97\%&66.84\%&10.00\%&56.31\%\\
\hline
\end{tabular}}

\end{small}
\end{table*}
 \begin{figure}[tb!]
\centerline{\includegraphics[width=\columnwidth]{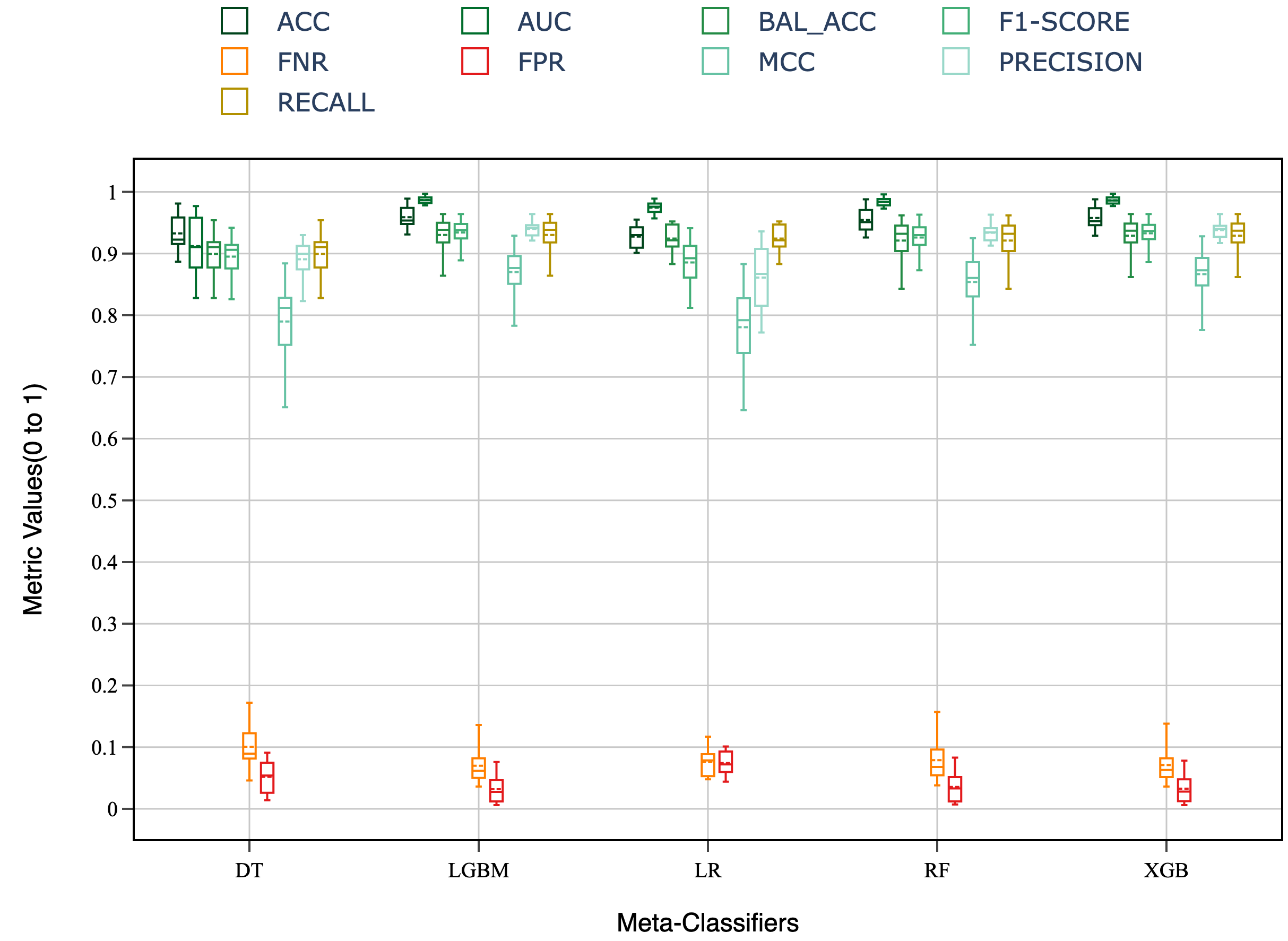}}
 \caption{Cross-validation performance of meta-classifiers (adversarial detectors) across all TTC models}
\label{tab:classifier-detection-overall}
 \end{figure}

  \begin{figure*}
\centerline{\includegraphics[width=\textwidth]{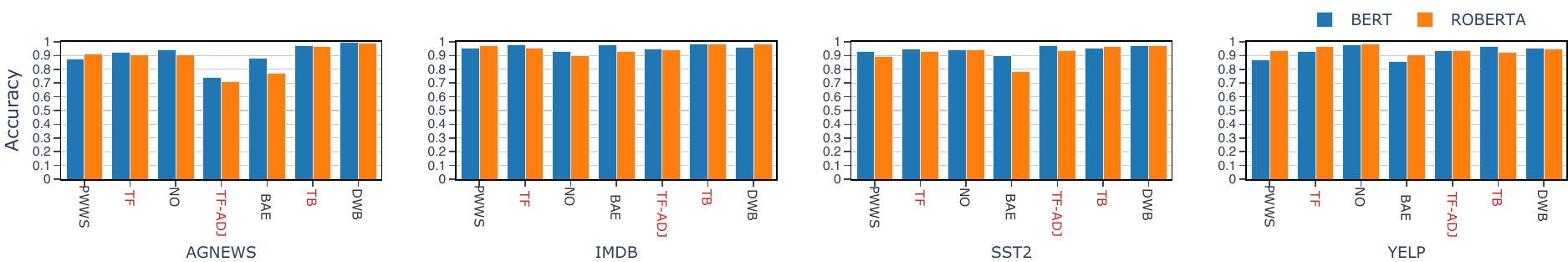}}
\caption{Comparison of $M_\mathrm{ad}$ accuracy on unseen data across different word substitution attacks (WSA). Red labels on the x-axis indicate attack types not used in training $M_\mathrm{ad}$. See section~\ref{Attacks_Studied} for details}
 \label{fig:classifier-perf-attackwise}
 \end{figure*}

\begin{figure*}[h!]
\centerline{\includegraphics[width=\textwidth]{Figs/fig5.pdf}}
\caption{Comparison of our method with WDR, EDIT (our), SHAP, and MLMDM detectors in detecting WSA}

\label{ComparisonAttackType}
 \end{figure*}
\subsection{Performance of TAD. }
\label{performanceTAD}
Fig. \ref{tab:classifier-detection-overall} shows the performance of meta-classifiers using boxplots illustration. The results reveal that LGBM and XGB consistently outperform other classifiers, with an average MCC of 0.87, F1-score above 93.5\%, and a Balanced Accuracy (BAL\_ACC) of 93.0\%. Notably, the LGBM model achieves the highest MCC of 0.929 for ROBERTA models trained on the YELP dataset. The LGBM classifier exhibits an FNR ranging from 3.6\% to 13.6\% and an FPR ranging from 0.6\% to 5\%. LR and DT classifiers show weaker performance, with average MCCs of 0.78 and 0.789, respectively. RFT performs well in some cases (average MCC of 0.854), but is outperformed by XGB and LGBM. 
Based on these findings, we select the LGBM classifier as our adversarial detector $M_\mathrm{ad}$.

\texttt{Comparison with SOTA}. 
Table \ref{Performance_Comparison_Small_Scale} presents the comparison of our approach with three SOTA methods. Our method demonstrates notable efficiency, reflected in its consistently lower feature extraction time (FET) of 0.085 seconds per example on average, contrasting with the computationally expensive SHAP (18 sec/sample), and moderately efficient WDR (6 sec/sample) and RDE (0.1375 sec/sample). Notably, FET varies across datasets and models, directly proportional to the median length of examples. For instance, IMDB exhibits a median length of 161 words, contributing to a higher FET, while SST-2, with a median length of 16 words, experiences a correspondingly lower FET.
Analyzing the performance of our adversarial detector ($M_\mathrm{ad}$), EDIT consistently exhibits strong results across all the datasets and models. For instance, for BERT model trained over AGNEWS dataset, EDIT outperforms competitors with the highest MCC of 0.77, surpassing WDR (0.76), SHAP (0.57), and RDE (0.42). 
Similar trends are observed in the ROBERTA model on the AGNEWS dataset, where EDIT consistently surpass other methods, showing substantial improvements across all the metrics.
For IMDB dataset (BERT), EDIT emerges as the top performer with the highest MCC (0.82), ACC (90.77\%) and F1-Score (90.76\%).
However, on the ROBERTA model for the IMDB dataset, WDR performed better than EDIT by 2\% across most of the metrics.
In the SST2 dataset, both BERT and ROBERTA models with EDIT consistently outperform competitors, achieving a remarkable MCC of 0.78 for the BERT model. In contrast, RDE shows sub-optimal performance with a negative MCC and high FPR and high FNR.
Finally, in the YELP dataset, EDIT dominates across both BERT and ROBERTA models with the highest MCC (0.87 and 0.89, respectively).
In summary, EDIT consistently outperforms WDR, SHAP, and RDE across the datasets and models, establishing itself as a robust and efficient method for detecting WSA.

We conducted further analysis (refer to Fig ~\ref{fig:classifier-perf-attackwise} and Fig~\ref{ComparisonAttackType}) to assess the performance of our EDIT detector module alongside other state-of-the-art (SOTA) detection methods.  
The results highlight EDIT robust performance, particularly with novel attacks such as A2T, TF, TB, and TF-ADJ (as seen by red color ins Fig~\ref{fig:classifier-perf-attackwise}, which were not used in training. Specifically, EDIT achieves a median accuracy of 95.5\% on the TF attack, 94.0\% on the TB attack, and 86.5\% on the TF-ADJ attack, indicating its strong generalizability in detecting previously unseen attacks.
For other attack types, EDIT consistently demonstrates superior accuracy: 83.5\% for BAE, 97.0\% for DWB, and 92.5\% for PWWS. When compared to other methods, EDIT outperforms SHAP, WDR, and MLMDM across all attack scenarios. For instance, SHAP records median accuracies of 65.5\% on BAE, 82.0\% on DWB, and 63.5\% on PWWS, whereas WDR achieves higher results with 85.0\% on BAE, 94.0\% on DWB, and 91.0\% on PWWS. MLMDM, however, falls behind with 46.5\% on both BAE and PWWS, and 53.5\% on DWB.
Analyzing the performance based on models, EDIT shows strong results with both BERT and ROBERTA, achieving median accuracies of 93.0\% and 91.5\%, respectively. This is superior to SHAP's 76.0\% and 65.0\%, WDR's consistent 91.0\% for both models, and MLMDM's lower accuracies of 52.0\% and 45.0\%.
Across different datasets, EDIT leads with high median accuracies: 91.0\% for AGNEWS, 94.0\% for IMDB, 89.5\% for SST2, and 95.5\% for YELP. In contrast, SHAP’s performance ranges from 68.5\% to 80.5\%, WDR's from 82.0\% to 94.0\%, and MLMDM records the lowest accuracies, ranging from 44.0\% to 61.5\%.
% Overall, the EDIT detector module demonstrates remarkable efficacy and generalizability, accurately classifying adversarial versus non-adversarial examples across various attacks, models, and datasets, significantly outperforming other methods.

 \subsection{Performance of TDT. }
 In this section, we present the performance of the TDT phase. Firstly, we outline the results of the identification module. Next, we examine the performance of the transformation and alerting sub-modules. Lastly, we discuss the overall performance of the TDT module.

 \begin{table*}[tb!]
\centering
\caption{Performance comparison with SOTA perturbed word identification methods with our EDIT identifier (descending order of performance)}
\begin{small}
\resizebox{\textwidth}{!}{\begin{tabular}{c|c|l|c|c|c|c|c|c|c}
\hline
\textbf{Dataset} & \textbf{Model} & \textbf{Methods} & \textbf{Bal\_ ACC} & \textbf{F1-Score} & \textbf{Recall} & \textbf{Precision} & \textbf{AUC} & \textbf{FPR} & \textbf{FNR} \\
\hline
\multirow{8}{*}{\textbf{AGNEWS}} & \multirow{4}{*}{\textbf{BERT}} 
& \textbf{EDIT} & \textbf{73.37\%} & \textbf{73.14\%} & \textbf{73.37\%} &\textbf{72.95\%} & \textbf{80.72\%} & 12.69\% &40.57\% \\
 &  & FreqScore & 70.91\% & 71.61\% & 70.91\% & 72.65\% & 77.95\% & \textbf{10.48\%} &  47.70\% \\
 &  & ExplainScore & 64.04\% & 58.01\% & 64.04\% & 59.97\% & 68.99\% & 39.31\% &\textbf{32.62\%} \\
 &  & ReplaceScore & 61.36\% & 57.45\% & 61.36\% & 58.32\% & 65.39\% & 34.91\% &42.37\% \\
\cline{2-10}
 & \multirow{4}{*}{\textbf{ROBERTA}} & \textbf{EDIT} & \textbf{69.79\%} & \textbf{70.44\%} & \textbf{69.79\%} & 71.33\% & \textbf{77.17\%} & 13.01\%& 47.41\% \\
  &  & FreqScore & 69.50\% & 70.40\% & 69.50\% & \textbf{71.82\%} & 76.87\% & \textbf{11.76\%} & 49.24\% \\
 &  & ExplainScore & 59.07\% & 54.55\% & 59.07\% & 56.97\% & 62.27\% & 44.65\% &\textbf{37.20\%} \\

 &  & ReplaceScore & 53.04\% & 49.85\% & 53.04\% & 52.34\% & 55.26\% & 46.80\% &47.12\% \\
\hline

\multirow{8}{*}{\textbf{IMDB}} & \multirow{4}{*}{\textbf{BERT}} & \textbf{EDIT} & \textbf{68.15\%} & \textbf{58.59\% }& \textbf{68.15\%} & \textbf{58.87\%} & \textbf{74.27\%} & \textbf{26.39\%} &\textbf{37.31\%} \\
  &  & ReplaceScore & 63.89\% & 53.99\% & 63.89\% & 56.17\% & 68.25\% & 31.88\% &40.35\% \\
&  & FreqScore & 61.54\% & 52.81\% & 61.54\% & 55.16\% & 66.65\% & 31.98\% &44.94\% \\
 &  & ExplainScore & 60.58\% & 53.37\% & 60.58\% & 55.05\% & 64.84\% & 29.30\%& 49.55\% \\

\cline{2-10}
 & \multirow{4}{*}{\textbf{ROBERTA}} & \textbf{EDIT} & \textbf{68.95\%} & \textbf{60.85\%} & \textbf{68.95\%} & 59.72\% & \textbf{75.48\%} & \textbf{20.02\%} &\textbf{42.07\%} \\
  &  & ExplainScore & 64.75\% & 54.83\% & 64.75\% & 56.23\% & 70.27\% & 28.09\% &42.40\% \\
 &  & ReplaceScore & 62.52\% & 52.43\% & 62.52\% & 55.04\% & 66.65\% & 31.67\% &43.29\% \\
  &  & FreqScore & 62.29\% & 56.38\% & 62.29\% & 56.19\% & 67.26\% & 21.04\% &54.39\% \\
\hline

\multirow{8}{*}{\textbf{SST2}} & \multirow{4}{*}{\textbf{BERT}} & \textbf{EDIT} & \textbf{73.09\%} & 69.88\% & \textbf{73.09\%} & \textbf{68.57\%} & \textbf{79.59\%} & \textbf{20.50\%} &\textbf{33.33\%} \\
 &  & ExplainScore & 68.91\% & 66.45\% & 68.91\% & 65.45\% & 74.88\% & 21.66\% &40.52\% \\
  &  & ReplaceScore & 66.41\% & 61.16\% & 66.41\% & 61.72\% & 72.83\% & 33.17\% &39.62\% \\
 &  & FreqScore & 64.21\% & 59.82\% & 64.21\% & 60.25\% & 69.05\% & 32.61\% &38.98\% \\

\cline{2-10}
 & \multirow{4}{*}{\textbf{ROBERTA}} & \textbf{EDIT} & \textbf{70.39\%} & \textbf{66.77\%} & \textbf{70.39\%} & \textbf{66.48\%} & \textbf{77.85\%} & 23.89\% &\textbf{38.71\%} \\
 &  & ExplainScore & 65.34\% & 60.89\% & 65.34\% & 61.29\% & 71.28\% & \textbf{22.14\%} &46.34\% \\
 &  & ReplaceScore & 65.11\% & 60.22\% & 65.11\% & 61.05\% & 70.49\% & 30.22\% &42.85\% \\
 &  & FreqScore & 63.28\% & 58.27\% & 63.28\% & 59.71\% & 68.93\% & 31.46\% &39.74\% \\
\hline
\multirow{8}{*}{\textbf{YELP}} & \multirow{4}{*}{\textbf{BERT}} & \textbf{EDIT} & \textbf{70.35\%} & \textbf{65.13\%} & \textbf{70.35\%} & 63.20\% & \textbf{76.73\%} & \textbf{15.82\%} &43.49\% \\
 &  & FreqScore & 67.93\% & 62.64\% & 67.93\% & \textbf{61.10\%} & 73.76\% & 17.35\% &46.79\% \\
 &  & ExplainScore & 66.89\% & 58.26\% & 66.89\% & 58.40\% & 71.96\% & 25.67\%& \textbf{40.55\%} \\
 &  & ReplaceScore & 64.55\% & 56.89\% & 64.55\% & 57.31\% & 68.38\% & 26.23\%& 44.68\% \\
\cline{2-10}
 & \multirow{4}{*}{\textbf{ROBERTA}}& EDIT & \textbf{75.71\%} & \textbf{68.99\%} & \textbf{75.71\%} & \textbf{66.55\%} & \textbf{83.48\%} & \textbf{16.13\%} &32.46\% \\
 &  & ExplainScore & 70.27\% & 58.80\% & 70.27\% & 59.99\% & 76.68\% & 31.00\%& \textbf{28.46\%} \\
 &  & FreqScore& 69.29\% & 64.83\% & 69.29\% & 63.11\% & 75.95\% & 16.24\% &45.18\% \\
 &  & ReplaceScore & 60.21\% & 54.90\% & 60.21\% & 55.59\% & 64.00\% & 26.92\%& 52.65\% \\
\hline
\end{tabular}}
\end{small}
\label{tab:perf_overall_identifier}
\end{table*}
 \begin{figure*}[h]
\centerline{\includegraphics[width=\textwidth]{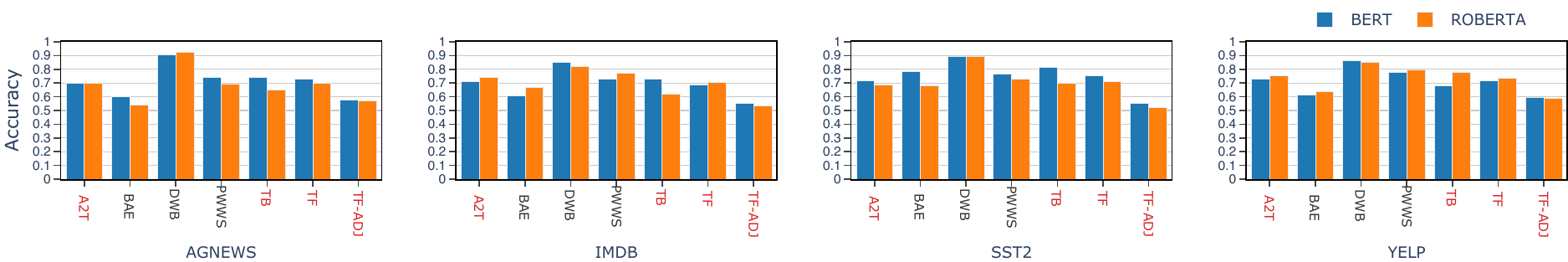}}
 \caption{Accuracy of the \new{identifier} on unseen word substitution attacks (WSA). Red labels on the x-axis indicate attacks not used in training. see section~\ref{Attacks_Studied} for details.}
 \label{fig:TDT-perf-identify}
 \end{figure*}
 
\begin{figure*}[htb!]
\centerline{\includegraphics[width=\textwidth]{Figs/Identifier_Results.pdf}}
 \caption{Comparison of our identifier EDIT, ExplainScore \cite{shen2023textshield}, ReplaceScore \cite{zhou2019learning} and FreqScore \cite{mozes2020frequency} }
 \label{fig:TDT-perf-identify-compare}
 \end{figure*}

%  \begin{figure*}[tb!]
% \centerline{\includegraphics[width=\linewidth, keepaspectratio]{Figs/fig6.pdf}}
%  \caption{Analysis of Transformation and Alerting module based on various Word Substitution Attacks, Datasets and Transformer models}
%  \label{fig:TA-perf-overall}
%  \end{figure*}

%  \begin{figure*}[tb!]
% \centerline{\includegraphics[width=\linewidth, keepaspectratio]{Figs/TDT.pdf}}
%  \caption{Overall Performance of TDT module based on various Word Substitution Attacks}
%  \label{fig:TDT-perf-overall-attacks}
%  \end{figure*}
%  \begin{figure*}[tb!]
% \centerline{\includegraphics[width=\linewidth, keepaspectratio]{Figs/fig6b.pdf}}
%  \caption{Overall Performance of TDT module based on various Word Substitution Attacks, Datasets and Transformer models}
%  \label{fig:TDT-perf-overall}
%  \end{figure*}

%  \begin{figure*}[tb!]
% \centerline{\includegraphics[width=\linewidth, keepaspectratio]{Figs/fig7.pdf}}
%  \caption{Time-taken by TDT module based on various Word Substitution Attacks, Datasets and Transformer models}
%  \label{fig:TDT-perf-overall-computation}
%  \end{figure*}
\subsubsection{Performance of Identification Module. }
Table~\ref{tab:perf_overall_identifier} reports the performance of the EDIT identifier across datasets and models. EDIT consistently achieves the highest Bal\_ACC. For example, on AGNEWS with BERT it reaches 73.37\%, surpassing FreqScore (70.91\%), ExplainScore (64.04\%), and ReplaceScore (61.36\%). This trend holds across combinations, showing EDIT balances the detection of perturbed and non-perturbed words better than baselines.  
In terms of F1-Score, EDIT also leads, reflecting a good precision–recall balance. On IMDB with ROBERTA, EDIT scores 60.85\%, compared to ExplainScore (54.83\%), ReplaceScore (52.43\%), and FreqScore (56.38\%). Similarly, on SST2 with BERT, EDIT attains the highest recall (73.09\%) with strong precision (68.57\%), whereas other methods trade off one for the other. EDIT achieves superior AUC values, such as 83.48\% on YELP with ROBERTA, indicating stronger overall classification. It also maintains lower error rates: on IMDB with BERT, EDIT reduces both FPR (26.39\%) and FNR (37.31\%) compared to all baselines. 

\new{\texttt{Generalization to unseen attacks.} We further assessed whether the identifier generalizes to perturbation strategies not seen during training. While trained only on BAE, DWB, and PWWS, it was tested against TF, TB, and TF\_ADJ. EDIT sustains strong performance: e.g., 71.4\% median accuracy on TextFooler and 72.8\% on TextBugger, despite no prior exposure. Its performance drops slightly on TF\_ADJ (56.1\%), but still exceeds competing methods. This shows that, like the adversarial detector, the identifier generalizes effectively to novel perturbations. The dip on TF\_ADJ stems from the subtle, context-integrated nature of its perturbations \cite{morris2020textattack}, making them harder to detect. }

\texttt{Model and dataset effects.} EDIT maintains robust performance across architectures, with slightly higher accuracy on BERT (72.7\%) than ROBERTA (69.9\%). Across datasets, it achieves the best results on YELP (73.4\%) and the lowest on AGNEWS (69.8\%), likely due to the challenges of multi-class classification. Overall, EDIT consistently outperforms ExplainScore, FreqScore, and ReplaceScore across domains, demonstrating adaptability and reliable perturbed word detection.

 \begin{figure*}[tb!]
\centerline{\includegraphics[width=\linewidth, keepaspectratio]{Figs/fig6.pdf}}
\caption{Analysis of the transformation and alerting module across different word substitution attacks, datasets, and transformer models}

 \label{fig:TA-perf-overall}
 \end{figure*}

 \begin{figure*}[tb!]
\centerline{\includegraphics[width=\linewidth, keepaspectratio]{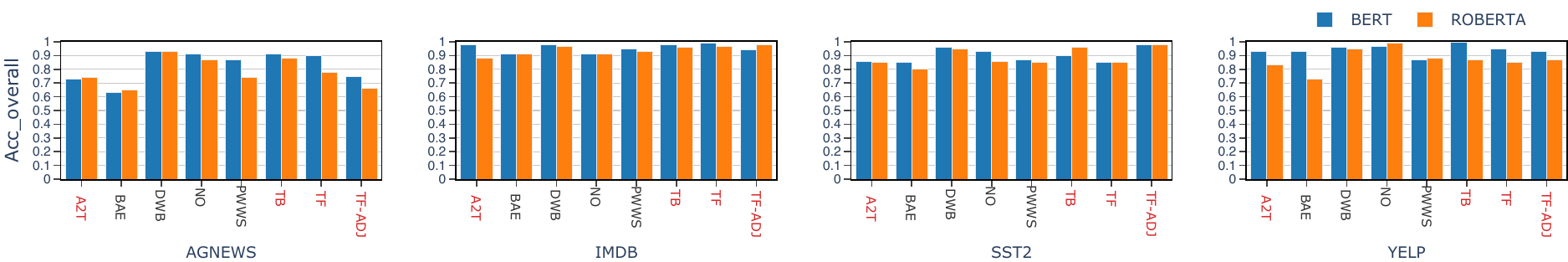}}
\caption{Overall performance of the TDT module across different word substitution attacks}

 \label{fig:TDT-perf-overall-attacks}
 \end{figure*}
 \begin{figure*}[tb!]
\centerline{\includegraphics[width=\linewidth, keepaspectratio]{Figs/fig6b.pdf}}
 \caption{Overall performance of TDT module based on various word substitution attacks, datasets and transformer models}
 \label{fig:TDT-perf-overall}
 \end{figure*}

 \begin{figure*}[tb!]
\centerline{\includegraphics[width=\linewidth, keepaspectratio]{Figs/fig7.pdf}}
 \caption{Time taken by the TDT module across different word substitution attacks, datasets, and transformer models}
 \label{fig:TDT-perf-overall-computation}
 \end{figure*}
\subsubsection{Performance of Transformation and Alerting. } 
 Fig \ref{fig:TDT-perf-overall} illustrates the effectiveness of the EDIT in transforming adversarial examples to non-adversarial ones, as indicated by $ACC\_transform$. It's evident that this effectiveness varies significantly across different attacks, with an average median accuracy of 88.60\% across all attacks. Our transformation module demonstrates the highest accuracies in defending against DWB attacks, achieving a median accuracy of 95.88\%, while BAE exhibits the lowest at 84.96\%. BAE low performance can be attributed from in contextually complex nature and identifier low performance (median accuracy 62.9\%) in detecting perturbed words. Also, it can be seen that for NO attack, our model has the lowest transformation accuracy 69.23\%. It means that if an example is detected incorrect as adversarial the true label is retained appoximately 70\% of the time. 
  Notably, EDIT achieves high transformation accuracies for both BERT and ROBERTA models with BERT surpassing ROBERTA (92.01\% vs. 88.89\%). 
 Among the datasets, YELP and IMDB boasts the highest median transformation accuracy at 96.70\% and 96.19\% respectively, whereas AGNEWS records the lowest at 80.45\%. The lower transformation accuracy observed in the AGNEWS dataset can be attributed to its multi-class classification nature, which introduces greater complexity in identifying and transforming adversarial examples compared to binary classification datasets.
 The ability to correctly alert analysts about potential adversarial examples shows considerable variability across attacks, models, and datasets as shown by ACC\_alert in Fig \ref{fig:TDT-perf-overall}. Overall EDIT generates the alerts for the security analyst with an average median accuracy of 87.5\%. With highest alert median accuracy ($>92\%$) for DWB, TB and TF-ADJ attacks , while BAE have the lowest (82.50\%). Both BERT and ROBERTA models exhibit similar median alerting accuracies, with BERT slightly outperforming ROBERTA (92\% vs. 89\%). Among the datasets, YELP and IMDB attained the highest median accuracies of 94\% and 93\% respectively.

\subsubsection{Overall Performance of EDIT Defense - TDT Phase. }

Figs ~\ref{fig:TDT-perf-overall-attacks} and~\ref{fig:TDT-perf-overall} provide an overview of the EDIT defense performance during the TDT phase. This encompasses how well the defense copes with adversarial attacks and a statistical breakdown across various attacks, models, and datasets, measured through Acc\_Det (detection accuracy), Acc\_without\_alert (accuracy without alert), and Acc\_overall (overall accuracy).
Figs~\ref{fig:TDT-perf-overall-attacks} and~\ref{fig:TDT-perf-overall}(a) demonstrate that, EDIT achieved an average overall accuracy (with alert) of 89\% and (without alert) of 84\%, respectively, in detecting attacks across datasets and models. Among the analyzed attacks, DWB, TB, TF-ADJ, and NO consistently demonstrated the highest Acc\_overall, with a median $>91\%$, indicating easier detection and mitigation of these attacks or preservation of original model performance. Attacks like A2T, BAE, and TF exhibited lower but competitive Acc\_overall (median $>85\%$), implying effective defense mechanism. Conversely, BAE had the lowest performance, with an Acc\_overall of 82.50\%, likely due to its subtle perturbations that are harder to detect.
Fig~\ref{fig:TDT-perf-overall}(b) shows that BERT outperformed RoBERTa consistently in Acc\_overall, indicating EDIT's higher efficacy for BERT models, possibly due to architectural differences.
Across datasets, as shown in Fig~\ref{fig:TDT-perf-overall}(c), IMDB consistently had the highest Acc\_overall, followed by YELP, SST2, and AGNEWS. The highest Acc\_overall was observed for BERT on the YELP dataset under the TB attack (1.00), while the lowest was for RoBERTa on AGNEWS under the BAE attack (0.65).
For both BERT and RoBERTa models trained on IMDB, our method performed exceptionally well, particularly under TB and TF attacks. For SST2, BERT excelled under TF-ADJ and DWB, while RoBERTa showed high resilience under TF-ADJ and TB. In the case of YELP, both BERT and RoBERTa exhibited high overall accuracies, with notable performance under specific attacks.
Furthermore, it is evident that Acc\_Det plays a crucial role in overall performance. A higher Acc\_Det ensures accurate detection of adversarial examples, enabling effective transformation into non-adversarial variants. Moreover, the importance of the alert module is evident, as Acc\_overall is significantly higher than Acc\_without\_alert.
Lastly, Fig \ref{fig:TDT-perf-overall-computation} shows the computational efficacy of our defenses. The analysis of the time taken by the EDIT defense mechanism reveals noteworthy patterns across different datasets, models, and attacks. The complexity of the attack plays a role in the transformation time, with more challenging attacks such as A2T and TF-ADJ requiring longer processing times compared to simpler attacks like NO or DWB. 
% For example, in the YELP dataset, the average transformation time for A2T and TF-ADJ attacks ranges from 8.50 to 12.69 seconds for BERT models and from 10.36 to 13.09 seconds for RoBERTa models, whereas for simpler attacks like NO attack and DWB, the times range from 3.28 to 6.02 seconds for RoBERTa models. 
Additionally, on average, the transformation process tends to be more time-consuming for RoBERTa models compared to BERT models, as evidenced by the significantly higher average times for RoBERTa across all datasets and attacks.
For instance, in the IMDB dataset, RoBERTa models exhibit notably longer transformation times compared to BERT models, with times ranging from 8.82 to 26.44 seconds for various attacks, while BERT models range from 6.62 to 16.49 seconds. 
Moreover, the transformation time varies across datasets, with more complex datasets such as IMDB and YELP consistently requiring longer transformation times compared to simpler datasets like SST2 and AGNEWS. 
% For instance, in the YELP dataset, the average transformation time ranges from 3.38 to 13.09 seconds for RoBERTa models and from 3.49 to 11.51 seconds for BERT models, whereas in the SST2 dataset, the times range from 1.68 to 4.59 seconds for RoBERTa models and from 1.83 to 4.15 seconds for BERT models.
One rational behind is that IMDB and YELP datasets, known for their longer sequences and higher average word lengths, contribute to the increased processing time.

\section{Analysis, Ablation Study and Discussion}
In this section, we delve into a comprehensive analysis and discussion of the key findings from the experimental results,

 \begin{table*}[htb!]
\caption{Comparison of explanation variance across training (Tr), testing (Te), and adversarial (Ad) distributions using attention maps ($\log{\sigma A_\mathrm{map}}$) and integrated gradients ($\log{\sigma I_{grad}}$). Metrics include Cohen's $d$ (effect size magnitude: low if $d < 0.8$, high if $d \geq 0.8$) and Bayes factor ($BF_{10}$, strength of evidence: higher values indicate stronger evidence, ``inf'' denotes extremely strong). Low effect implies stable explanations across distributions, while high effect indicates significant variation.}

\label{RQ1_Chapter6}
\centering
\begin{small}
\resizebox{\textwidth}{!}{\begin{tabular}{c|c|c|ccc|ccc}
\hline
\multirow{2}{*}{\textbf{Dataset}} & \multirow{2}{*}{\textbf{Model}} &\multirow{2}{*}{\textbf{Distributions}} & \multicolumn{3}{c}{Attention Map ($\log\sigma{A_\mathrm{map}}$)} & \multicolumn{3}{c}{\textbf{Integrated Gradients ($\log\sigma{I_{grad}}$)}} \\
\cline{4-6} \cline{7-9}
 &  &  & \textbf{d} & \textbf{Effect Size} & \textbf{$BF_{10}$}& \textbf{d} & \textbf{Effect Size} & \textbf{$BF_{10}$} \\
\hline
\multirow{6}{*}{\textbf{AGNEWS}} & \multirow{3}{*}{\textbf{BERT}} & Tr-Te & 0.038 & low & 1.338 & 0.129 & low & 1.23E+15 \\
 &  & Tr-Ad & 1.962 & high & inf & 3.571 & high & inf \\
 &  & Te-Ad & 2.329 & high & inf & 2.179 & high & inf \\
  \cline{2-9}
 & \multirow{3}{*}{\textbf{ROBERTA}} & Tr-Te & 0.088 & low & 2.242E+06 & 0.116 & low & 6.957E+10 \\
 &  & Tr-Ad & 1.985 & high & inf & 2.319 & high & inf \\
 &  & Te-Ad & 2.112 & high & inf & 1.852 & high & inf \\
 \hline
 \multirow{6}{*}{\textbf{IMDB}} & \multirow{3}{*}{\textbf{BERT}} & Tr-Te & 0.078 & low & 3.113E+13 & 0.147 & low & 2.08E+53 \\
 &  & Tr-Ad & 1.610 & high & inf & 3.835 & high & inf \\
 &  & Te-Ad & 1.523 & high & inf & 3.269 & high & inf \\
 \cline{2-9}
 & \multirow{3}{*}{\textbf{ROBERTA}} & Tr-Te & 0.040 & low & 169.115 & 0.071 & low & 1.107E+11 \\
 &  & Tr-Ad & 1.618 & high & inf & 2.579 & high & inf \\
 &  & Te-Ad & 1.564 & high & inf & 2.426 & high & inf \\
 \hline
 \multirow{6}{*}{\textbf{SST2}} &  \multirow{3}{*}{\textbf{BERT}}& Tr-Te & 0.105 & low & 122.059 & 0.224 & low & 4.201E+11 \\
 &  & Tr-Ad & 1.399 & high & inf & 1.910 & high & inf \\
 &  & Te-Ad & 1.385 & high & inf & 1.497 & high & inf \\
 \cline{2-9}
 &  \multirow{3}{*}{\textbf{ROBERTA}} & Tr-Te & 0.061 & low & 0.564 & 0.148 & low & 1.013E+05 \\
 &  & Tr-Ad & 1.282 & high & inf & 1.884 & high & inf \\
 &  & Te-Ad & 1.327 & high & inf & 1.581 & high & inf \\
 \hline
 \multirow{6}{*}{\textbf{YELP}} &  \multirow{3}{*}{\textbf{BERT}} & Tr-Te & 0.033 & low & 2871.167 & 0.071 & low & 1.10E+22 \\
 &  & Tr-Ad & 2.293 & high & inf & 3.692 & high & inf \\
 &&Te-Ad&2.341&high&inf&3.276&	high&	inf\\
  \cline{2-9}
 &  \multirow{3}{*}{\textbf{ROBERTA}} & Tr-Te &	0.0514	&low&	21120000000	&0.0297&	low	&100.298\\
&  & Tr-Ad &	2.053&	high	&inf	&4.445&	high&	inf\\
 &  & Te-Ad &	2.165&	high&	inf&	4.142&	high&	inf\\
 \hline
\end{tabular}}
\end{small}
\end{table*}

% \begin{figure*}[htb!]
% \centering\includegraphics[width=0.6\linewidth]{Figs/analysiskde.png}
% \caption{\updated{
% KDE of \texttt{gradient\_std} and \texttt{explain\_std} across clean and adversarial examples. 
% Although some overlap exists, a notable statistical separation supports effective detection.
% }}
% \label{figure_analysis_kde}
% \end{figure*}

% \begin{figure*}[tb!]
% \includegraphics[width=\textwidth, height=0.8\textwidth, keepaspectratio]{Figs/Significance_yelp.png}
% \caption{{Token perturbation significance and impact on YELP dataset. Adversarial tokens tend to be rare and semantically manipulative.}}
% \label{AttentionMap}
% \end{figure*}
% \begin{figure*}[htb!]
% \centering\includegraphics[width=\linewidth]{Figs/ExplainAnalysisIdentification.png}
% \caption{\updated{Explanation attribution divergence. Adversarial tokens receive disproportionately high importance.}}
% \label{fig:explain_analysis}
% \end{figure*}

\updated{\subsection{Theoretical and Empirical Justification for Explainability Features. }}
\updated{Explainability methods such as attention maps and integrated gradients (IG) identify tokens critical to model predictions. For clean inputs, these importance scores align with semantically relevant tokens. Adversarial examples, however, deliberately shift model focus to irrelevant tokens, causing a measurable saliency misalignment.}
\updated{Formally, let \( x \) be a clean input and \( x' \) its adversarial counterpart. If \( \phi(x) \) denotes an explainability mapping (e.g., IG vector or attention weights), we expect \( \phi(x) \approx \phi(x') \) under benign changes. However, adversarial perturbations intentionally cause the divergence, where \( \epsilon \) is a small tolerance bound. This divergence is precisely what enables explainability metrics to differentiate between clean and adversarial inputs. 
\begin{equation}
   \|\phi(x) - \phi(x')\|_2 \gg \epsilon,
\end{equation}
Prior work also shows adversarial examples exploit brittle, “non-robust” features detectable via saliency shifts \cite{ilyas2019adversarial}, and robust models yield more interpretable explanations \cite{etmann2019connection}.}
\textbf{Empirical Analysis.} We analyze explainability patterns across datasets (\texttt{YELP}, \texttt{AG\_NEWS}, \texttt{IMDB}, \texttt{SST2}) with BERT and RoBERTa classifiers. Table~\ref{RQ1_Chapter6} reports significant statistical differences between benign and adversarial inputs in attention map and IG distributions, quantified by large Cohen’s d effect sizes and overwhelming Bayes factors \(BF_{10}\). No meaningful differences arise between training and test sets, confirming stability for genuine inputs \new{(Refer to Supplementary material for more details)}.
% Kernel density estimates (Fig~\ref{figure_analysis_kde}) reveal clear separation in gradient and explanation variance between clean and adversarial samples, consistent across models and datasets.}
% At the token level (Fig~\ref{AttentionMap}), adversarial edits often replace tokens with rare or syntactically noisy alternatives, maintaining semantic coherence but shifting model focus. Explanation attribution distributions (Fig~\ref{fig:explain_analysis}) further show adversarial tokens receive disproportionately high importance, offering a reliable forensic signal.
% In summary, explainability features robustly expose adversarial perturbations by detecting internal representation drift, underpinning methods like \textsc{EdIT} that leverage these signals for effective detection and robustness evaluation.}
\subsection{Impact of Training Data Diversity on \texorpdfstring{$M_{\text{ad}}$}{Mad} Performance. }
\label{unknown_attacks}
 \begin{figure}[tb]
    \centering
    \includegraphics[width=\linewidth, keepaspectratio]{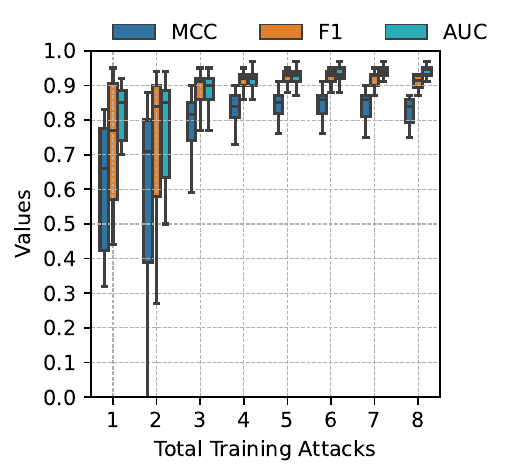}
    \caption{Effect of attack diversity on detector performance}
    \label{Chapter6_RQ2_Diversity}
\end{figure}

The effectiveness of the adversarial detector ($M_\mathrm{ad}$) relies on its training data, which is crucial for learning the distinctive features of AEs and distinguishing them from clean inputs. To understand the relationship between performance and attack diversity in the training data, we conducted an ablation study. We evaluated the model's performance with varying numbers of adversarial attacks (0 to 7) as shown in Fig~\ref{Chapter6_RQ2_Diversity}.
Training the detector exclusively on normal samples (0 attacks) establishes a baseline using an Auto-Encoder based anomaly detector. Fig~\ref{Chapter6_RQ2_Diversity} shows that increasing the diversity of attacks in the training data generally improves both median and mean performance metrics. This suggests that a more diverse set of attacks enhances the adversarial detector's effectiveness.
However, there are diminishing returns; for instance, moving from 6 to 7 attacks does not yield a significant performance boost, indicating that the effect plateaus after a certain point. This highlights the importance of diverse attacks in training while balancing the cost and benefit.
Additionally, exposing $M_\mathrm{ad}$ to various attacks during training enhances its ability to spot new, unseen attacks. Performance leveling off after about four attack types suggests adaptability to new attack methods, a crucial feature for an effective adversarial detector in real-world scenarios.

\subsection{Resiliency to Adaptive Attacks. }
\begin{figure*}[tb!]
    \centering
    \includegraphics[width=\textwidth]{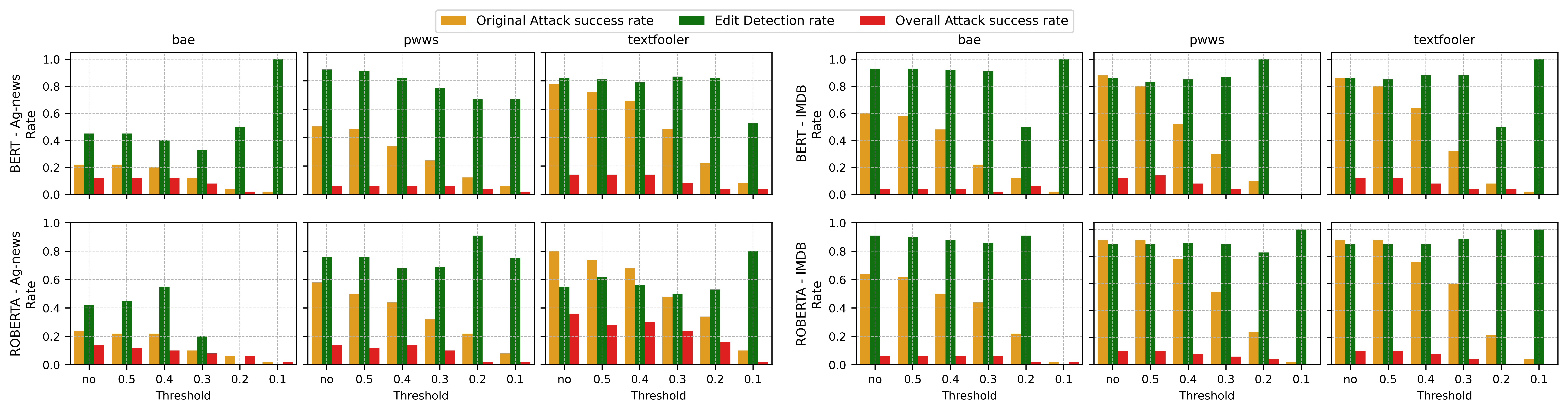}
 \caption{Performance of baseline (``no'') versus adaptive threshold attacks across models and datasets, showing original attack success rate (OASR), edit detection rate (EDR), and overall attack success rate (OverASR)}

    \label{fig:overall_adaptive_attacks}
\end{figure*}

The proactive and transformative nature of \textsc{EdIT} poses a substantial obstacle to adaptive adversarial attacks, especially those relying on model feedback to craft subtle perturbations. By enforcing internal feature similarity and transforming inputs, \textsc{EdIT} disrupts the attacker's optimization loop, making successful evasion far more difficult.

\updated{To evaluate \textsc{EdIT} under a strong adaptive threat model, we assume the attacker has full knowledge of the defense and seeks to evade detection. We modify three attack methods: TextFooler~\cite{jin2019bert}, PWWS~\cite{ren2019generating}, and BAE~\cite{garg2020bae}—to incorporate a novel \texttt{EditFeatureConstraint}, which filters token substitutions based on their impact on internal representations such as attention patterns and integrated gradients.
We implement this constraint using the TextAttack framework, allowing only edits that maintain high similarity to the original input in terms of attention and gradient features. Given a perturbed input $x'$, we compute a composite similarity score:
}
\begin{equation}
d(x, x') = w_a \cdot (1 - \cos(\mathbf{a}, \mathbf{a'})) + w_g \cdot (1 - \cos(\mathbf{g}, \mathbf{g'})),
\end{equation}
\updated{
\noindent where \(\mathbf{a}, \mathbf{g}\) and \(\mathbf{a'}, \mathbf{g'}\) represent the attention and gradient vectors of the original and perturbed inputs, respectively. The weights \(w_a\) and \(w_g\) control the relative importance of attention and gradient similarities, with values set to 0.6 and 0.4 in our experiments. A perturbation is accepted only if the composite similarity score \(d(x, x')\) is below a threshold \(\tau\), which governs the strictness of the edit similarity constraint and ranges from 0.1 to 0.5.
\textsc{EdIT} combines attention and gradient similarities into a single unified score and flags inputs as ``edited'' if this score falls below \(\tau\). Typically, benign paraphrases preserve these internal features, whereas adversarial edits cause measurable deviations that \textsc{EdIT} can detect.}
\updated{
Fig~\ref{fig:overall_adaptive_attacks} summarizes results across models and datasets. We report:
(i) Original Attack Success Rate (OASR) – the proportion of adversarial examples that induce misclassification;
(ii) Edit Detection Rate (EDR) – the fraction of adversarial inputs flagged by \textsc{EdIT}; and
(iii) Overall Attack Success Rate (OverASR) – the proportion of examples that both fool the model and evade detection.
Our analysis reveals that across both models BERT and RoBERTa, and for both AG-News and IMDB datasets, the OverASR under adaptive attacks is significantly reduced—frequently below 0.14 and often dropping to 0 at stricter thresholds. This demonstrates that once the attacker must balance both misclassification and stealth, the attack space becomes highly constrained.
Additionally, while some attacks (e.g., TextFooler on RoBERTa IMDB) achieve high OASR (up to 0.92), these same attacks are almost entirely detected (EDR $\approx$ 0.89–1.0), leading to OverASR $\approx$ 0.1 or lower. Conversely, attacks that evade detection (low EDR) typically fail to alter model predictions, confirming a fundamental trade-off between effectiveness and stealth in adversarial design. Moreover,  non-adaptive (baseline) attacks, which are unconstrained by edit similarity, achieve marginally higher OverASRs in some cases (e.g., 0.36 for RoBERTa on AG-News with TextFooler). However, their higher visibility (lower EDR) reinforces the fact that \textsc{EdIT} is especially effective at detecting aggressive or poorly disguised perturbations. Finally, \textsc{EdIT} maintains strong detection capabilities across model architectures and domains. RoBERTa, while slightly more vulnerable under baseline conditions, shows similar OverASR suppression under adaptive constraints. IMDB’s longer sequences also seem to enhance feature-based detection, particularly for BAE and PWWS attacks.}

\updated{These results affirm two key properties of \textsc{EdIT}: (1) constraining internal representation drift shrinks the adversarial space, and (2) edit-aware detection enforces a trade-off attackers must choose between evasion and impact. Together, these make \textsc{EdIT} a resilient and broadly applicable defense.}

\subsection{\updated{Transferability and Robustness under Attacker and Defender Model Mismatch. }}
\updated{To evaluate the robustness of \textsc{Edit} in more realistic threat scenarios, we investigate its performance under a model mismatch between the attacker and defender. Specifically, we consider two settings: (1) adversarial examples (AEs) are generated using RoBERTa, while the \textsc{Edit} detector is trained on and deployed with BERT; and (2) AEs are generated using BERT, while the detector is trained on and deployed with RoBERTa. 
We evaluate three attack algorithms, TextFooler, PWWS, and BAE on the \textsc{AG News} dataset. For each setup, we report the \textit{transferability rate} (i.e., fraction of AEs that successfully transfer to the defender model), the \textit{detection rate} of \textsc{Edit}, and the \textit{residual attack success rate} after detection.
Fig~\ref{fig:attacker_defender_heatmaps} summarizes cross‐model attack transferability, detection rates, and ultimate success. Although BAE attains the highest raw transferability (0.42 for BERT $\to$ RoBERTa, 0.28 for RoBERTa $\to$ BERT), \textsc{Edit} flags over half of these examples (0.49/0.67), yielding final success rates of only 0.06 and 0.03, respectively. PWWS and TextFooler transfer less effectively ($\leq$0.14/0.09) but are detected at $\ge$0.64, resulting in overall attack success between 0.01 and 0.04. These results underscore \textsc{Edit}’s ability to generalize across architectures and constrain adversarial efficacy to low, single‐digit percentages, an essential characteristic for robust real‐world deployment.
}

\begin{figure*}[ht]
  \centering
  \includegraphics[width=\textwidth]{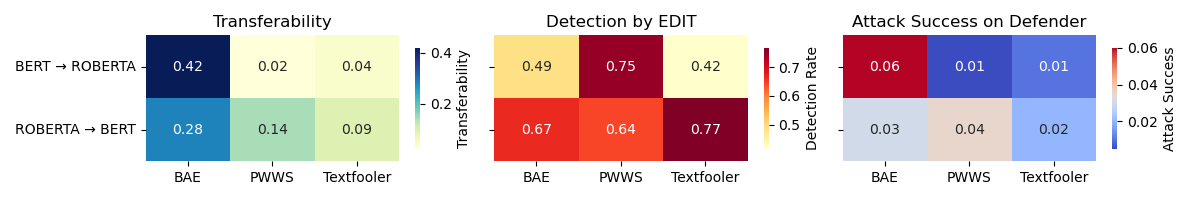}
\caption{Heatmaps comparing attack performance across different attacker–defender model pairs. The y-axis represents attacker→defender pairs (e.g., BERT→RoBERTa), and the x-axis represents attack methods}

  \label{fig:attacker_defender_heatmaps}
\end{figure*}
\begin{table*}[tb!]
\centering
\small
\caption{Ablation study across different \textit{opt\_score} formulations (Flags 0–4). Best values per column are in bold. Fluency (PPL$\uparrow$) is computed as the inverted, normalized perplexity score so that higher values indicate greater fluency, consistent with the other metrics}

\resizebox{\textwidth}{!}{\begin{tabular}{lccccc}
\hline
\textbf{Flag} & \textbf{Recovered Label (Acc)} & \textbf{$\Delta$ PCS} & \textbf{Semantic (Token-level)} & \textbf{Fluency (PPL$\uparrow$)} & \textbf{Semantic (Sentence-level)} \\
\hline
0 & \textbf{0.800 $\pm$ 0.163} & \textbf{0.844 $\pm$ 0.132} & \textbf{0.613 $\pm$ 0.095} & 0.652 $\pm$ 0.380 & \textbf{0.855 $\pm$ 0.072} \\
1 & 0.781 $\pm$ 0.173 & 0.819 $\pm$ 0.153 & 0.612 $\pm$ 0.093 & \textbf{0.656 $\pm$ 0.376} & 0.854 $\pm$ 0.073 \\
2 & 0.762 $\pm$ 0.164 & 0.800 $\pm$ 0.154 & 0.604 $\pm$ 0.094 & 0.642 $\pm$ 0.377 & 0.846 $\pm$ 0.075 \\
3 & 0.762 $\pm$ 0.189 & 0.806 $\pm$ 0.180 & 0.604 $\pm$ 0.091 & 0.636 $\pm$ 0.392 & 0.851 $\pm$ 0.069 \\
4 & 0.794 $\pm$ 0.176 & 0.831 $\pm$ 0.165 & 0.611 $\pm$ 0.093 & 0.647 $\pm$ 0.385 & \textbf{0.855 $\pm$ 0.070} \\
\hline
\end{tabular}}
\label{tab:ablation}
\end{table*}

% \subsection{Scalability and Efficiency}
% The EDIT framework typically requires an average of 6.94 seconds (with observed durations ranging from a minimum of 1.68 secs to a maximum of 26.44 seconds, as shown in Fig~\ref{fig:TDT-perf-overall-computation}
% })
% to detect and transform an adversarial example into a non-adversarial one. This duration is primarily attributed to the transformation process, which involves replacing perturbed words and can be time-consuming. The complexity of this transformation mechanism is represented by $O(n*d)$, where $n$ denotes the number of words to replace and $d$ indicates the possible replacements for each word. While parallelization techniques can expedite the transformation by utilizing multiple GPUs, the availability of computational resources may limit the degree of parallelization achievable. Despite these time constraints, the transformation step remains crucial for converting adversarial examples into non-adversarial ones and enhancing the model's robustness. Future research endeavors could concentrate on optimizing the transformation process and investigating efficient parallelization methods to mitigate the time required.
\subsection{Scalability and Efficiency. }
The \textsc{EDIT} framework requires on average 6.94 seconds (range: 1.68--26.44s, Fig.~\ref{fig:TDT-perf-overall-computation}) to detect and transform an adversarial example, with the transformation step dominating this cost. Its complexity is $O(n \times d)$, where $n$ is the number of words replaced and $d$ the candidate pool size.  

\new{\noindent\textbf{Detection vs. Transformation.} Runtime consists of two phases. The detection phase, which identifies adversarial examples, is relatively fast (1--2s on average; see Table~\ref{Performance_Comparison_Small_Scale}), whereas the transformation step accounts for most of the latency. Parallelization (e.g., multi-GPU deployment) can reduce this cost depending on resources. Importantly, the real-time applicability of \textsc{EDIT} refers to the detection phase, while the slower transformation phase is primarily used in offline workflows.}  

\new{\noindent\textbf{Deployment Modes.} EDIT supports two modes that balance latency against analyst workload.}  

\begin{enumerate}
    \item \new{In the \textit{real-time detect-and-alert} setting (1--2s), the system flags adversarial inputs for immediate analyst action. This is suitable for scenarios such as enterprise phishing defense, customer support chatbots in banking or healthcare, or moderation of live social media streams where rapid alerts are critical.}  
    \item \new{In the \textit{detect--transform} mode ($\sim$6.94s), the full pipeline is executed and is more appropriate when real-time response is less important but robustness is required, for example in automated moderation of YouTube or Twitter comments, filtering adversarial prompts in enterprise AI assistants, or forensic analysis and model hardening in security labs.}  
\end{enumerate}

\new{\noindent\textbf{Alert Volumes.} To estimate analyst workload, we evaluated 500 clean and 500 TextFooler adversarial inputs on IMDB and AG-News with BERT. Alerts were extremely rare ($\leq$0.8\%), while overall correctness exceeded 90\% across datasets (Table~\ref{tab:alert_stats}). False detections on clean inputs (up to 10.2\%) were internally resolved without escalation, confirming a low analyst burden.}  

\begin{table*}[h]
\centering
\caption{Alert and correctness statistics on 500 clean and 500 adversarial inputs for IMDB and AG-News with BERT (percentages are relative to the total)}

\label{tab:alert_stats}
\begin{tabular}{llccc}
\toprule
\textbf{Dataset} & \textbf{Category} & \textbf{Analyst Alerts (\%)} & Detected as Adv \textbf{(\%)} & \textbf{Accuracy (\%)} \\
\midrule
\textbf{IMDB}     & Clean        & 0.2 & 10.2 & 91.2 \\
\textbf{IMDB}     & Adversarial  & 0.0 & 82.4 & 95.4 \\
\textbf{AGNews}  & Clean        & 0.4 & 4.0  & 96.0 \\
\textbf{AGNews}  & Adversarial  & 0.8 & 91.0 & 90.0 \\
\bottomrule
\end{tabular}
\end{table*}

\new{\noindent\textbf{Analyst Workflow.} Each alert is delivered as a structured threat-intelligence report (see Supplementary Material Fig 5), which includes the flagged input, highlighted perturbations, ranked candidate replacements, confidence deltas, and a concise outcome message. Analysts then follow a triage process: (a) dismiss benign alerts, (b) validate or accept automated transformations if correct, or (c) escalate unresolved cases for deeper review. This ensures alerts are interpretable, actionable, and low in volume.}  

{\noindent\textbf{Future Directions.} Optimizing transformation through GPU acceleration, efficient synonym generation, or candidate pruning will further reduce latency and extend EDIT’s applicability in real-time settings.}

\subsection{Ablation Study on Transformation .}
\new{Experiments were conducted on the IMDB and AG\_News datasets with BERT and RoBERTa classifiers under two adversarial attack methods (TextFooler and PWWS). For each configuration, we generated 20 adversarial examples.  }

\subsubsection{Ablation Study on Opt\_Score. }
\new{To assess the contribution of each component in the proposed \textit{opt\_score} (see Algorithm \ref{algo2}), we performed an ablation study. The \textit{opt\_score} is defined as a linear combination of four complementary signals: (1) $\Delta PCS$, capturing model confidence shifts; (2) token-level semantic similarity (\textit{simi\_score}); (3) sentence-level semantic similarity (\textit{sub\_simi\_score}); and (4) frequency-based plausibility (\textit{freq\_score}).  
We evaluated five variants of \textit{opt\_score}: \textbf{Flag 0} (full combination), \textbf{Flag 1} (\textit{freq\_score} only), \textbf{Flag 2} (\textit{simi\_score} only), \textbf{Flag 3} (\textit{sub\_simi\_score} only), and \textbf{Flag 4} ($\Delta PCS$ only). The results (Table~\ref{tab:ablation}) show that the full combination (\textbf{Flag 0}) achieves balanced performance across all metrics, while individual components emphasize one dimension but underperform elsewhere. For instance, \textbf{Flag 1} slightly improves fluency, and \textbf{Flag 4} maintains sentence-level similarity; however, both fail to sustain high accuracy. This demonstrates that combining all components ensures robustness by jointly addressing accuracy recovery, semantic preservation, and fluency.  }

\subsubsection{Evaluating the Linguistic Quality of Transformed Text. }
\label{sec:linguistic_quality}
\new{While the transformation module restores classifier predictions, it is equally important to verify the linguistic quality of the output. We evaluated this using two complementary approaches: (1) automated metrics and (2) qualitative assessment via GPT-4o. Fig~\ref{fig:C2_linguistic} summarizes the findings.  }

\new{\noindent\textbf{Automated Metrics.}  
Following prior work \cite{schmidtova-etal-2024-automatic-metrics}, we used three measures: (1) \textbf{Token-level semantic similarity:} BERTScore F1 \cite{zhangbertscore}, which correlates with human judgments of meaning preservation. (2) \textbf{Sentence-level semantic similarity:} cosine similarity of Sentence-BERT embeddings \cite{jing-etal-2023-multi}, capturing overall semantic overlap. (3) \textbf{Fluency:} normalized perplexity under a pretrained LM \cite{yang-jin-2023-attractive}, reflecting grammaticality and naturalness. Fig.~\ref{fig:C2_linguistic} (a) shows token-level similarity around 0.6, indicating moderate lexical overlap with the original text. Sentence-level similarity remains consistently high ($>0.8$ median), confirming that meaning is preserved even when surface forms differ. Fluency is more variable but maintains a high mean, suggesting that most outputs remain grammatical and natural.}

\noindent\textbf{Qualitative Large Language Model (LLM) Assessment.}  
\new{We further employed GPT-4o as a proxy evaluator, as LLM-based judgments align well with human annotations \cite{chiang-lee-2023-large}. For each instance, GPT-4o was presented with the original, adversarial, and transformed text and asked to rate \textit{semantic preservation}, \textit{grammatical correctness}, and \textit{fluency} on a 1--5 scale, with justifications (see Supplementary Material for the prompt). Two authors manually validated 10 randomly selected evaluations. Fig.~\ref{fig:C2_linguistic} (b) shows that semantic preservation and grammaticality consistently score around 3--4, confirming that outputs remain faithful in meaning and grammatically sound. Fluency scores are slightly lower, reflecting occasional awkward phrasing, though still acceptable overall. Variability across metrics is modest, indicating stable judgments across examples. Importantly, since \textsc{EDIT} only has access to the adversarial text, its transformation is constrained to the minimal edits needed to restore the original label. As a result, certain adversarial linguistic disruptions cannot be fully removed, which explains the residual imperfections in fluency. Overall, automated metrics highlight strong sentence-level semantic preservation, while GPT-4o ratings surface residual fluency issues, together confirming that transformations are accurate in meaning though occasionally less natural in style.}}

\begin{figure}[ht]
  \centering
  \includegraphics[width=\linewidth]{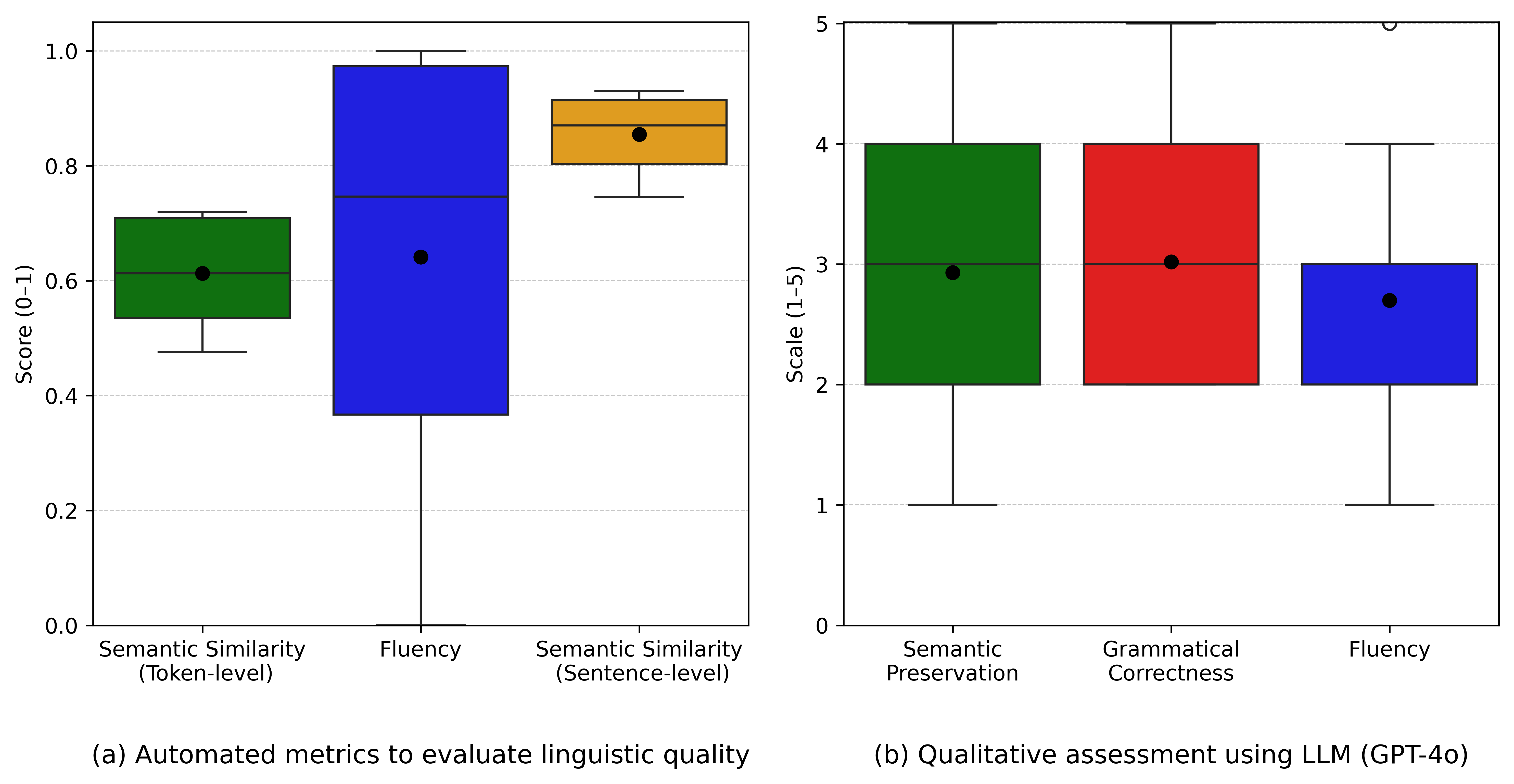}
\caption{Evaluation of the linguistic quality of transformed text using automated metrics (left) and qualitative assessment with GPT-4o (right)}
  \label{fig:C2_linguistic}
\end{figure}

\updated{\subsection{Threats to Validity and Future Work}}
Our approach has several limitations that guide directions for future work:
   \begin{enumerate}
       \item  The method currently targets transformer-based models exclusively. Future research will extend it to other architectures such as CNNs and LSTMs, and incorporate explainability techniques beyond attention mechanisms to improve generalizability. 
       \item Certain adversarial attacks, including TF-ADJ and BAE, remain difficult to detect, particularly on datasets like AGNEWS. We plan to explore additional features and advanced techniques to enhance detection and robustness against a broader spectrum of attacks.
       \item  Experiments were conducted on offline adversarial examples (AEs). We will evaluate the framework in real-time settings, especially for score-based attacks, to assess effectiveness and efficiency in dynamic environments.
       \item The practical utility of generated security logs has not been fully assessed. Future work includes utility evaluations and gathering feedback from security analysts to improve usability and real-world applicability.
       \item  The transformation step relies on pre-trained embeddings and synonym resources (e.g., WordNet, GloVe, BERT-MLM), which may encode social or domain-specific biases and reduce accuracy in some contexts. Although mitigated by combining multiple substitution sources and model feedback, bias remains a concern. Investigating bias mitigation and domain adaptation is an important avenue for future work.
   \end{enumerate}

\updated{Addressing these limitations will enhance our framework’s robustness across diverse models, datasets, real-time scenarios, and practical security applications, ultimately strengthening its defense against adversarial attacks.
}
\section{Conclusion}
\label{sec:conclusion}
The EDIT framework offers a robust solution for mitigating Word Substitution Attacks on transformer-based models. Our approach encompasses the development of an adversarial detector and techniques for identifying and transforming AEs, yielding significant results. The framework achieves a median detection Matthews Correlation Coefficient of 0.846 and an overall median accuracy of 92.98\% for the Test-Time Detection and AE Transformation  method. Furthermore, it demonstrates strong performance in accurately identifying perturbations (median recall of 70\%), effectively transforming AEs (median accuracy of 91\%), and correctly involving human intervention (median accuracy of 89\%). Our future work encompasses three key areas: (i) extending the framework to cater to non-transformer-based models, (ii) enhancing the utility of generated logs by involving security analysts, and (iii) evaluating the framework's performance in real-time scenarios. These advancements will augment the framework's capabilities and expand its efficacy in defending against various adversarial attacks.

\bibliographystyle{IEEEtran}
\bibliography{reference}
\vskip -2\baselineskip plus -1fil
\begin{IEEEbiography}[{\includegraphics[width=1in,height=1.25in,clip,keepaspectratio]{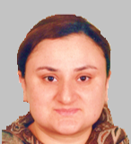}}]{Bushra Sabir} is a Research Scientist at CSIRO’s Data61, specializing in Adversarial Machine Learning and Secure AI. With over 14 years of experience spanning AI, cybersecurity, and secure software systems, she contributes to projects at the intersection of AI, networking, and security-critical infrastructure. Her PhD, awarded with a Dean’s Commendation, focused on defending machine learning models against advanced threats. Prior to Data61, she completed a postdoctoral fellowship on adversarial attacks in 6G and secure code generation, and worked as a Senior Lecturer and software engineer. She has published in top venues including IEEE TDSC, NAACL, MSR, and ACM Computing Surveys.
\end{IEEEbiography}
\vskip -2\baselineskip plus -1fil
\begin{IEEEbiography}[{\includegraphics[width=1in,height=1.25in,clip]{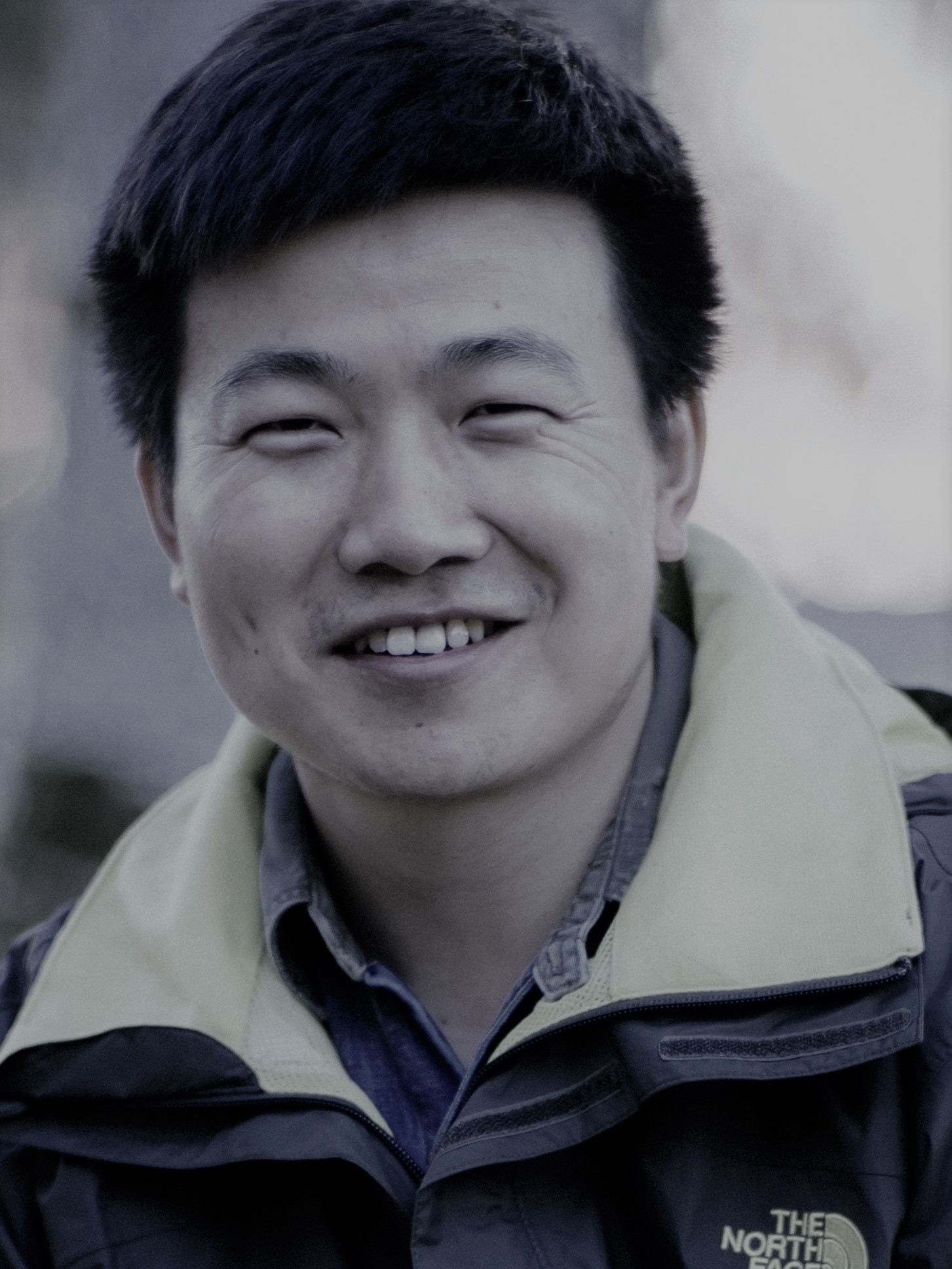}}]{Dr Yansong Gao} is a Lecturer (Assistant Professor equivalent) at the University of Western Australia. Formerly, he was a tenured Research Scientist and later a Senior Research Scientist at CSIRO's Data61. A Senior Member of IEEE, he serves as an Associate Editor for IEEE TIFS (since Oct 2024) and IEEE TNNLS (since Jan 2024). He has received Best Paper Awards at Usenix Security 2024 and AsiaCCS 2023, and an Outstanding PC Member Award at AsiaCCS 2024. He holds an M.Sc. from UESTC and a Ph.D. from the University of Adelaide.
\end{IEEEbiography}
\vskip -2\baselineskip plus -1fil
\begin{IEEEbiography}[{\includegraphics[width=1in,height=1.25in,clip,keepaspectratio]{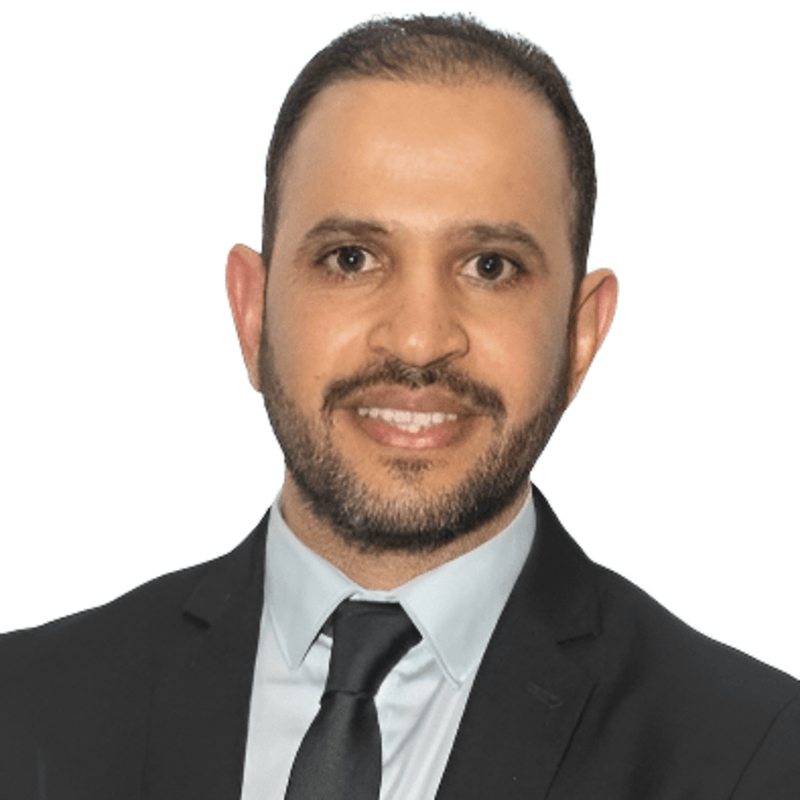}}]{Dr Alsharif Abuadbba} is Team Leader of Distributed Systems Security and Senior Research Scientist at CSIRO’s Data61. With 14+ years of experience, he leads national cybersecurity projects and has developed award-winning AI-driven solutions like Smartshield and TAPE. He has published 70+ papers (IEEE S\&P, NDSS, Usenix Security), serves as Associate Editor for IEEE TIFS, and supervises PhD students.
A PhD graduate from RMIT and recipient of CSIRO’s Julius Career Award and the 2024 Australia AI Innovator—Cybersecurity Lead Award, he also co-founded the Eureka-winning startup EyeCura and contributes to national cybersecurity and AI safety standards.
\end{IEEEbiography}
\vskip -2\baselineskip plus -1fil
\begin{IEEEbiography}[{\includegraphics[width=1in,height=1.25in,clip,keepaspectratio]{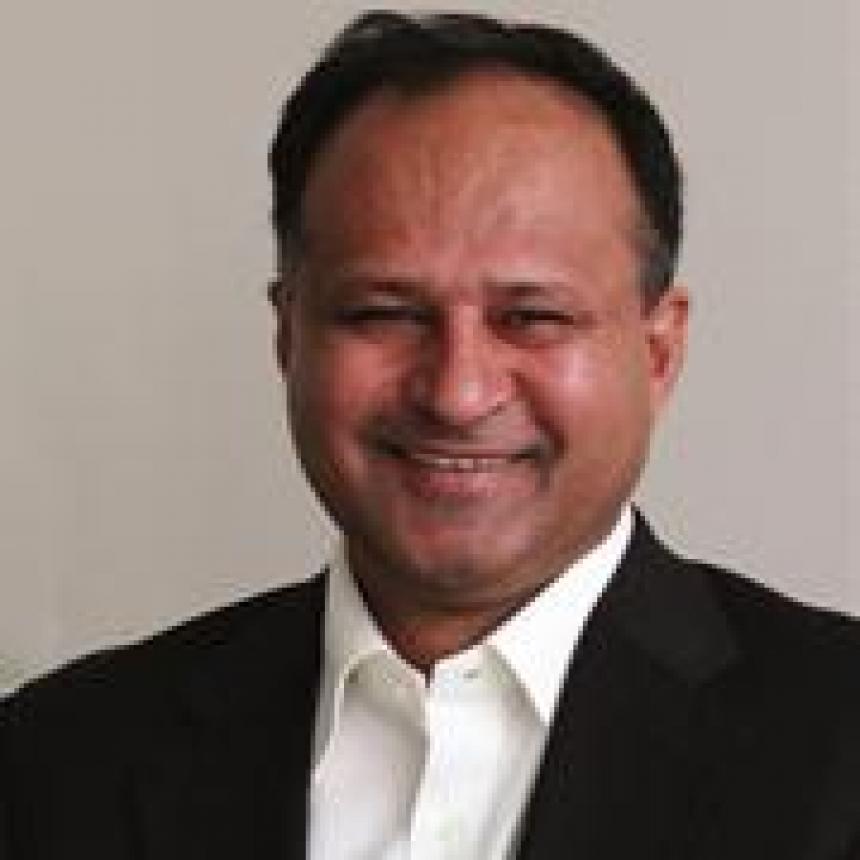}}]{M. Ali Babar}
is a leading researcher in Software Engineering with over 18 years of experience, focusing on secure software systems for emerging technologies such as Cloud, Edge, IoT, and Big Data. His work lies at the intersection of Software Engineering, AI (ML/NLP), and Cybersecurity, employing both qualitative and quantitative empirical methods.
He has published 320+ peer-reviewed papers with 19293 citations (h-index 69, as of Jul 2025), placing him among the top Software Engineering researchers in Australia/New Zealand. His research has received multiple awards, including Most Influential Paper at ASE 2014. At the University of Adelaide, he led a successful bid for the \$140M Cyber Security Cooperative Research Centre (CSCRC), where he heads the “Platform and Architecture for Cyber Security as a Service” theme. His research is conducted in close collaboration with major industry and government partners, including DST Group, ATO, Cisco, and AC3.
\url{https://researchers.adelaide.edu.au/profile/ali.babar#my-research}.
\end{IEEEbiography}
\input{appendix}
\end{document}

%% file: appendix.tex
\appendix
\section{Appendix}
We provide additional results and analysis in this section.
\begin{table}[!htb]
\caption{Baseline Datasets detail [Here Tr\_N and Te\_N refers to number of training and testing samples]}
\label{Chapter6_dataset}
\centering
\begin{small}
\resizebox{\linewidth}{!}{\begin{tabular}
{|>{\raggedright\arraybackslash}m{25mm}|m{25mm}|m{7mm}|m{10mm}|m{7mm}|m{7mm}|}
\hline
\textbf{Dataset} & \textbf{Task} & \textbf{Classes} & \textbf{Median Length} & \textbf{Tr\_N} & \textbf{Te\_N} \\ \hline
IMDB \cite{maas2011learning} & Movie Review & 2 & 161 & 25k & 25k \\ \hline
AG-News \cite{zhang2015character} & Headline Topic & 4 & 44 & 120k & 7.6k  \\ \hline
SST-2 \cite{socher2013recursive} & Movie Review & 2 & 16 & 2.7k & 1821  \\ \hline
YELP \cite{zhang2015character} & Restaurant Review & 2 & 152 & 100k & 38k \\ \hline
\end{tabular}}
\end{small}
\end{table}

\begin{table*}[htb!]
\centering
\caption{Adversarial Dataset Details with number of adversarial samples}
\label{adversarialdataset_Chapter6}
\begin{small}
\resizebox{\linewidth}{!}{\begin{tabular}{|c|c|c|c|c|c|c|c|c|}
\hline
\textbf{Dataset} & \textbf{Model} & \textbf{DWB} & \textbf{TF} & \textbf{BAE} & \textbf{PWWS} & \textbf{TF-ADJ} & \textbf{TB} & \textbf{Total} \\ \hline
\multirow{2}{*}{\textbf{IMDB}} & \textbf{BERT} & 714 & 9164 & 5959 & 9082 & 1415 & 854 & 27188 \\ \cline{2-9}
& \textbf{ROBERTA} & 514 & 10212 & 7289 & 10338 & 870 & 863 & 30086 \\ \hline
\multirow{2}{*}{\textbf{SST2}} & \textbf{BERT} & 728 & 1636 & 1039 & 1491 & 80 & 471 & 5445 \\ \cline{2-9}
& \textbf{ROBERTA} & 765 & 1650 & 1062 & 1501 & 61 & 409 & 5448 \\ \hline
\multirow{2}{*}{\textbf{AGNEWS}} & \textbf{BERT} & 576 & 5895 & 1034 & 4140 & 343 & 483 & 12471 \\ \cline{2-9}
& \textbf{ROBERTA} & 541 & 6179 & 1190 & 4720 & 367 & 497 & 13494 \\ \hline
\multirow{2}{*}{\textbf{YELP}} & \textbf{BERT} & 697 & 4588 & 2613 & 4572 & 489 & 822 & 13781 \\
\cline{2-9}
& \textbf{ROBERTA} & 1509 & 1734 & 2301 & 1655 & 106 & 1200 & 8505 \\ \hline
\end{tabular}}
\end{small}
\end{table*}

\begin{figure*}[htb!]
\centering\includegraphics[width=\linewidth]{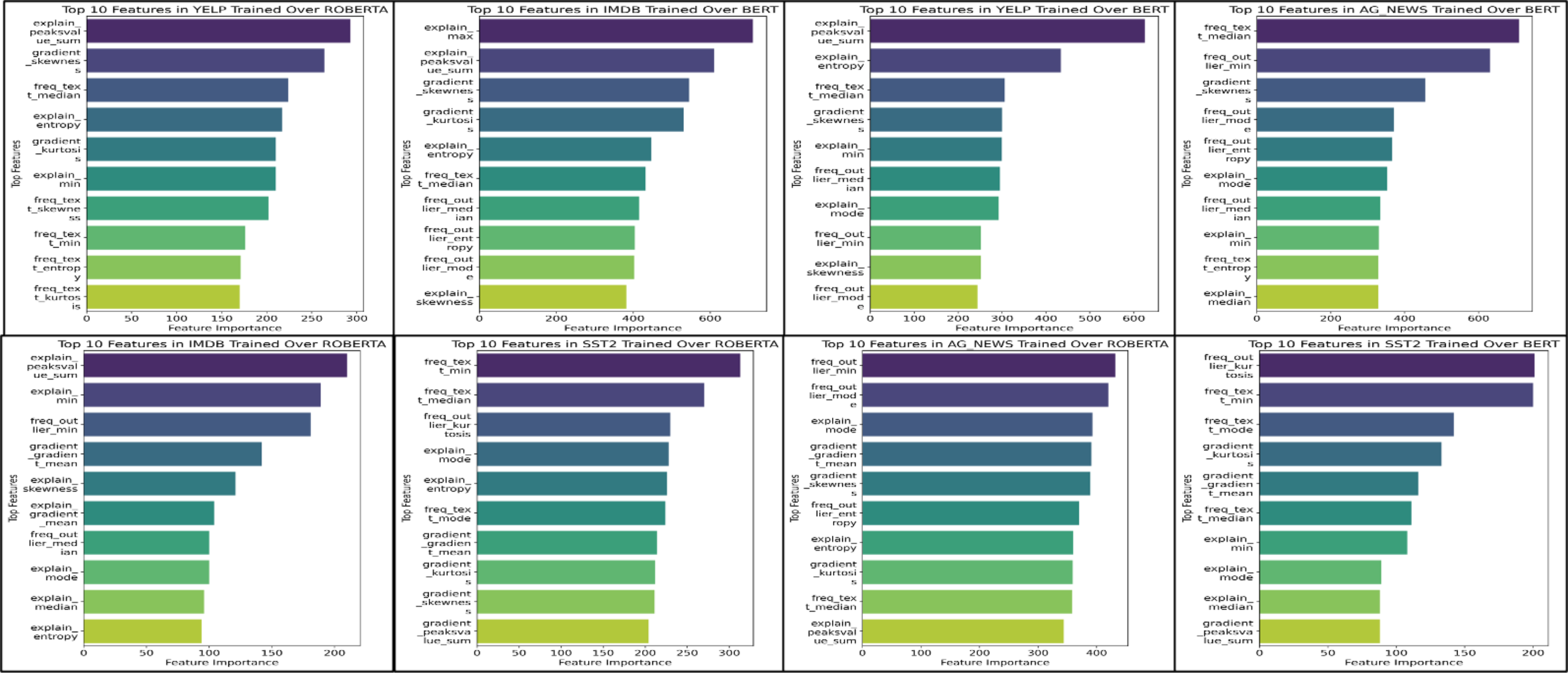}
\caption{Feature Importance Across Dataset and $T\_m$ models}
\label{FeaturesImp}
\end{figure*}

\textbf{Explainability Analysis and Effectiveness} 
\begin{figure*}[htb!]
\centering\includegraphics[width=\linewidth]{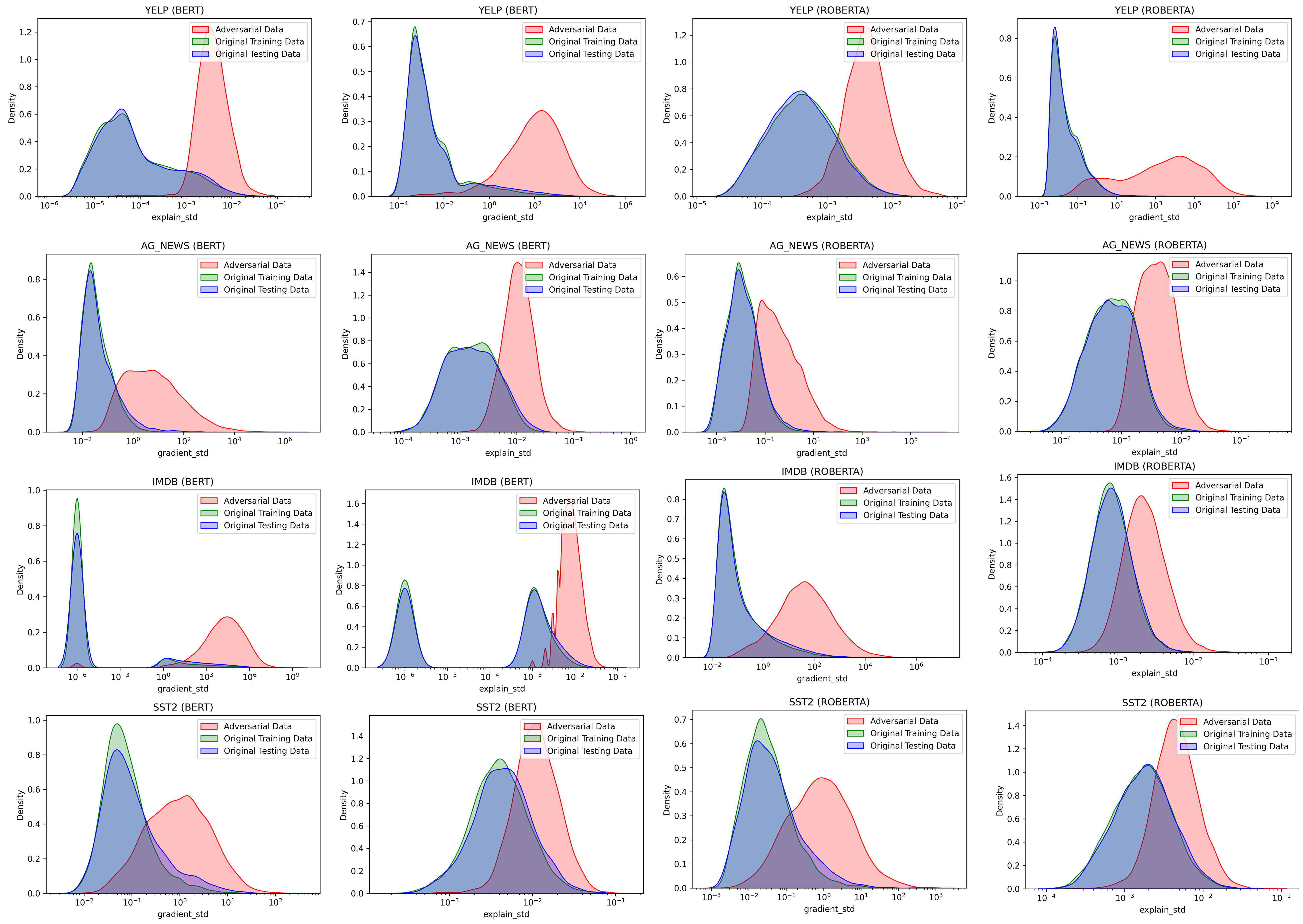}
\caption{
KDE of \texttt{gradient\_std} and \texttt{explain\_std} across clean and adversarial examples. 
Although some overlap exists, a notable statistical separation supports effective detection.
}
\label{figure_analysis_kde}
\end{figure*}
\begin{figure*}[tb!]
\centering
\includegraphics[width=\linewidth]{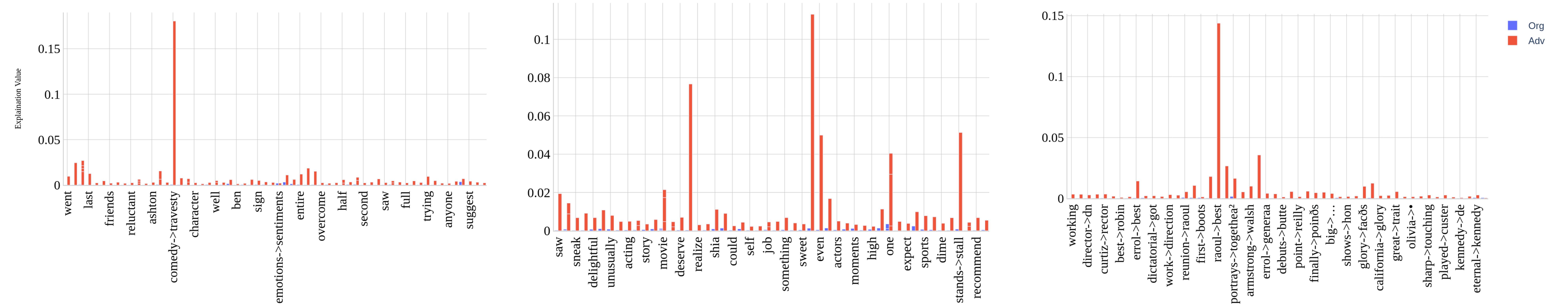}
\caption{{Explanation attribution divergence. 
Adversarial tokens receive disproportionately high importance compared to clean tokens.}}
\label{fig:explain_analysis}
\end{figure*}

% \begin{figure*}[htb]
% \includegraphics[width=0.8\textwidth]{Figs/Significance_yelp.png}
% \caption{{Token perturbation significance and impact on YELP dataset. Adversarial tokens tend to be rare and semantically manipulative.}}
% \label{AttentionMap}
% \end{figure*}

% \begin{figure*}[htb!]
% \centerline{\includegraphics[width=0.8\linewidth]{Figs/threatreportAdv1.png}}
% \caption{Sample Threat Intelligence Report.}
% \label{threatreportAdv1}
% \end{figure*}
\begin{figure*}[htb]
    \centering
    \includegraphics[height=0.3\textheight, width=\textwidth,keepaspectratio]{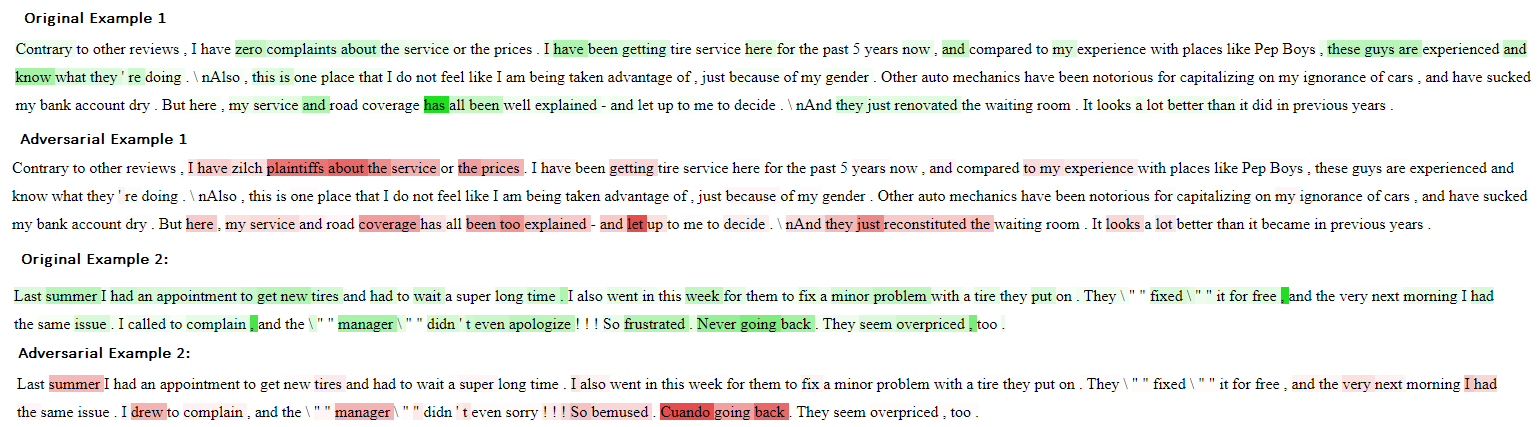}
    \caption{Token perturbation significance and impact on YELP dataset. Adversarial tokens tend to be rare and semantically manipulative.}
    \label{AttentionMap}
\end{figure*}

\begin{figure*}[htb]
    \centering
    \includegraphics[height=0.6\textheight,width=\textwidth, keepaspectratio]{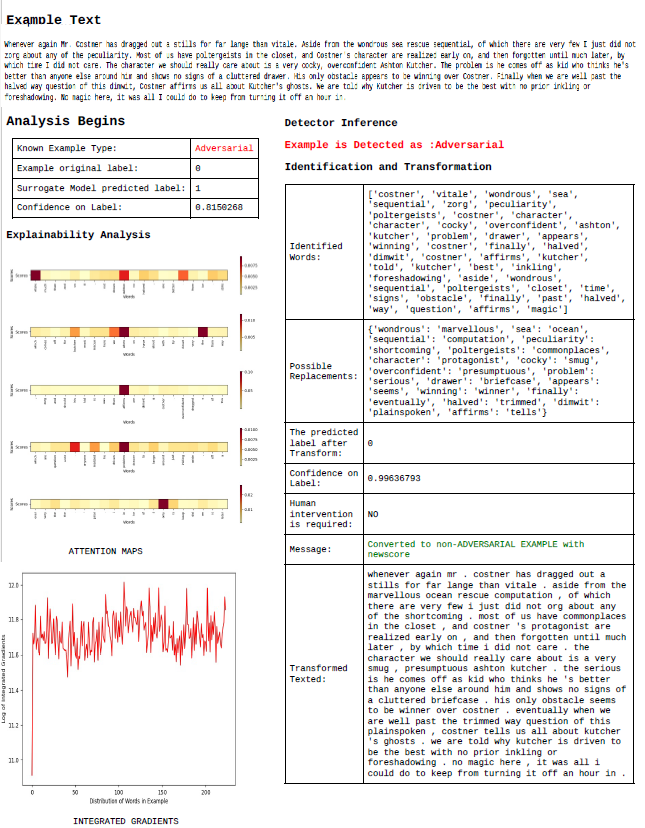}
    \caption{Sample Threat Intelligence Report.}
    \label{threatreportAdv1}
\end{figure*}

Kernel density estimates (Fig~\ref{figure_analysis_kde}) reveal clear separation in gradient and explanation variance between clean and adversarial samples, consistent across models and datasets.
At the token level (Figure~\ref{AttentionMap}), adversarial edits often replace tokens with rare or syntactically noisy alternatives, maintaining semantic coherence but shifting model focus. Explanation attribution distributions (Figure~\ref{fig:explain_analysis}) further show adversarial tokens receive disproportionately high importance, offering a reliable forensic signal.  
In summary, explainability features robustly expose adversarial perturbations by detecting internal representation drift, underpinning methods like \textsc{EdIT} that leverage these signals for effective detection and robustness evaluation.

\textbf{Handling Overfitting}
To counteract potential overfitting, we implemented several strategies: an attack-based stratified dataset split to diversify exposure during training, an attack-agnostic approach to ensure versatility across various adversarial scenarios, and 10-fold cross-validation to guarantee the model's ability to generalize across different data samples. We used the Matthews Correlation Coefficient (MCC) as the model selection criterion due to its resilience against imbalanced datasets and its comprehensive consideration of all classes. Additionally, Bayesian optimization was employed to fine-tune the models. The dataset was divided into training (70\%) and testing (30\%) sets, with MCC as the primary performance criterion. Stratified sampling with 10-fold cross-validation ensured balanced class distribution in each fold. Classifier performance was evaluated using MCC, Balanced Accuracy (BalACC), F1 Score, Area Under the Curve (AUC), False Positive Rate (FPR), and False Negative Rate (FNR). Furthermore, we analyzed the selected $M_{ad}$'s accuracy in detecting specific attacks.

\textbf{Impact of Detection Error Propagation}
Given the significant reliance of the Test-Time Detection and AE Transformation (TDT) module on the output of $M_\mathrm{ad}$, it is reasonable to expect that any detection errors will affect the performance of the TDT module.
Two types of errors need consideration in this context: (i)  FNR, where an AE is incorrectly identified as benign, and (ii) FPR, indicating a normal example being flagged as adversarial. It's noteworthy that FNR does not transfer to the TDT module, illustrating the error rate in the EDIT framework. As our results for TDT module demonstrates that the FNR of our trained detector is consistently below 10\%, with the exception of the Roberta model trained on AGNEWS.
Conversely, FPR does propagate to the TDT module. As observed, our TDT module exhibits a median ACC\_Transform of 69.23\% in the absence of an attack scenario, however, EDIT still attained an ACC\_overall of 91\% in this case. This implies that the FPR error is not directly transferred to the TDT module, which suggests that TDT is equipped to manage it by either bypassing the transformation process or seeking human intervention to address the FPR.

\textbf{Handling Overfitting}
To counteract potential overfitting, we implemented several strategies: an attack-based stratified dataset split to diversify exposure during training, an attack-agnostic approach to ensure versatility across various adversarial scenarios, and 10-fold cross-validation to guarantee the model's ability to generalize across different data samples. We used the Matthews Correlation Coefficient (MCC) as the model selection criterion due to its resilience against imbalanced datasets and its comprehensive consideration of all classes. Additionally, Bayesian optimization was employed to fine-tune the models. The dataset was divided into training (70\%) and testing (30\%) sets, with MCC as the primary performance criterion. Stratified sampling with 10-fold cross-validation ensured balanced class distribution in each fold. Classifier performance was evaluated using MCC, Balanced Accuracy (BalACC), F1 Score, Area Under the Curve (AUC), False Positive Rate (FPR), and False Negative Rate (FNR). Furthermore, we analyzed the selected $M_{ad}$'s accuracy in detecting specific attacks.

\textbf{Impact of Detection Error Propagation}
Given the significant reliance of the Test-Time Detection and AE Transformation (TDT) module on the output of $M_\mathrm{ad}$, it is reasonable to expect that any detection errors will affect the performance of the TDT module.
Two types of errors need consideration in this context: (i)  FNR, where an AE is incorrectly identified as benign, and (ii) FPR, indicating a normal example being flagged as adversarial. It's noteworthy that FNR does not transfer to the TDT module, illustrating the error rate in the EDIT framework. As our results for TDT module demonstrates that the FNR of our trained detector is consistently below 10\%, with the exception of the Roberta model trained on AGNEWS.
Conversely, FPR does propagate to the TDT module. As observed, our TDT module exhibits a median ACC\_Transform of 69.23\% in the absence of an attack scenario, however, EDIT still attained an ACC\_overall of 91\% in this case. This implies that the FPR error is not directly transferred to the TDT module, which suggests that TDT is equipped to manage it by either bypassing the transformation process or seeking human intervention to address the FPR.

\begin{tcolorbox}[
  colback=white, 
  colframe=black, 
  coltext=black,
  left=4pt, 
  right=4pt, 
  top=4pt, 
  bottom=4pt,
  width=\textwidth,
  title={\textbf{Linguistic Quality Evaluation for Adversarial Text Transformation}}
]
\footnotesize
\begin{verbatim}
You are a linguistic evaluator. You are given three versions of a sentence with their labels:

Original (label = {original_label}): {original}  
Adversarial (label = {adv_label}): {adversarial}  
Transformed (label = {trans_label}): {transformed}  

The transformation is produced by the EDIT framework, which detects adversarial examples  
and minimally edits the adversarial text to restore the original label.  

Please evaluate the transformed text on the following criteria (scale 1–5, where 1 = poor, 5 = excellent):  

1. Semantic preservation  
2. Grammatical correctness  
3. Fluency  

Provide your answer in strict JSON format:  
{
   "semantic_preservation": score,  
   "grammar": score,  
   "fluency": score,  
   "justification": "short explanation"  
}
\end{verbatim}
\end{tcolorbox}